%% file: main.tex
\documentclass{article} % For LaTeX2e
\usepackage{iclr2026_conference,times}
\input{math_commands.tex}

\usepackage{booktabs} 
\usepackage{multirow} 
\usepackage{caption}  
\usepackage{graphicx} 
\usepackage{xcolor}
\definecolor{lightgray}{rgb}{0.8,0.8,0.8} 
\usepackage{graphicx} % For \includegraphics
\usepackage{subcaption} 

\usepackage{calc} 
\usepackage{rotating} % For \rotatebox
\newlength{\imagecolwidth}
\newlength{\rowlabelcolwidth}
\newcommand{\Centering}{\centering}
\usepackage{float}
\usepackage{afterpage}

\usepackage{hyperref}
\usepackage{url}
\usepackage{amsmath} 
\usepackage{cleveref}

\makeatletter
\def\@fnsymbol#1{\ensuremath{\ifcase#1\or \dagger\or \ddagger\or \mathsection\or \mathparagraph\or \|\or **\or \dagger\dagger \or \ddagger\ddagger \else\@ctrerr\fi}}
\makeatother

\iclrfinalcopy
\renewcommand{\headrulewidth}{0pt}
\newcommand{\name}{BRIDGE}

\title{\name\ - Building Reinforcement-Learning Depth-to-Image Data Generation Engine for Monocular Depth Estimation}

\author {
    Dingning Liu\textsuperscript{\rm 1,2}
    Haoyu Guo\textsuperscript{\rm 1}
    Jingyi Zhou\textsuperscript{\rm 1}
    Tong He\textsuperscript{\rm 1}\thanks{Corresponding author.} \\
\textsuperscript{\rm 1}Shanghai Artificial Intelligence Laboratory  
\textsuperscript{\rm 2}Fudan University
}

\begin{document}

\maketitle

\begin{figure}[htbp]
    \centering
\includegraphics[width=\linewidth]{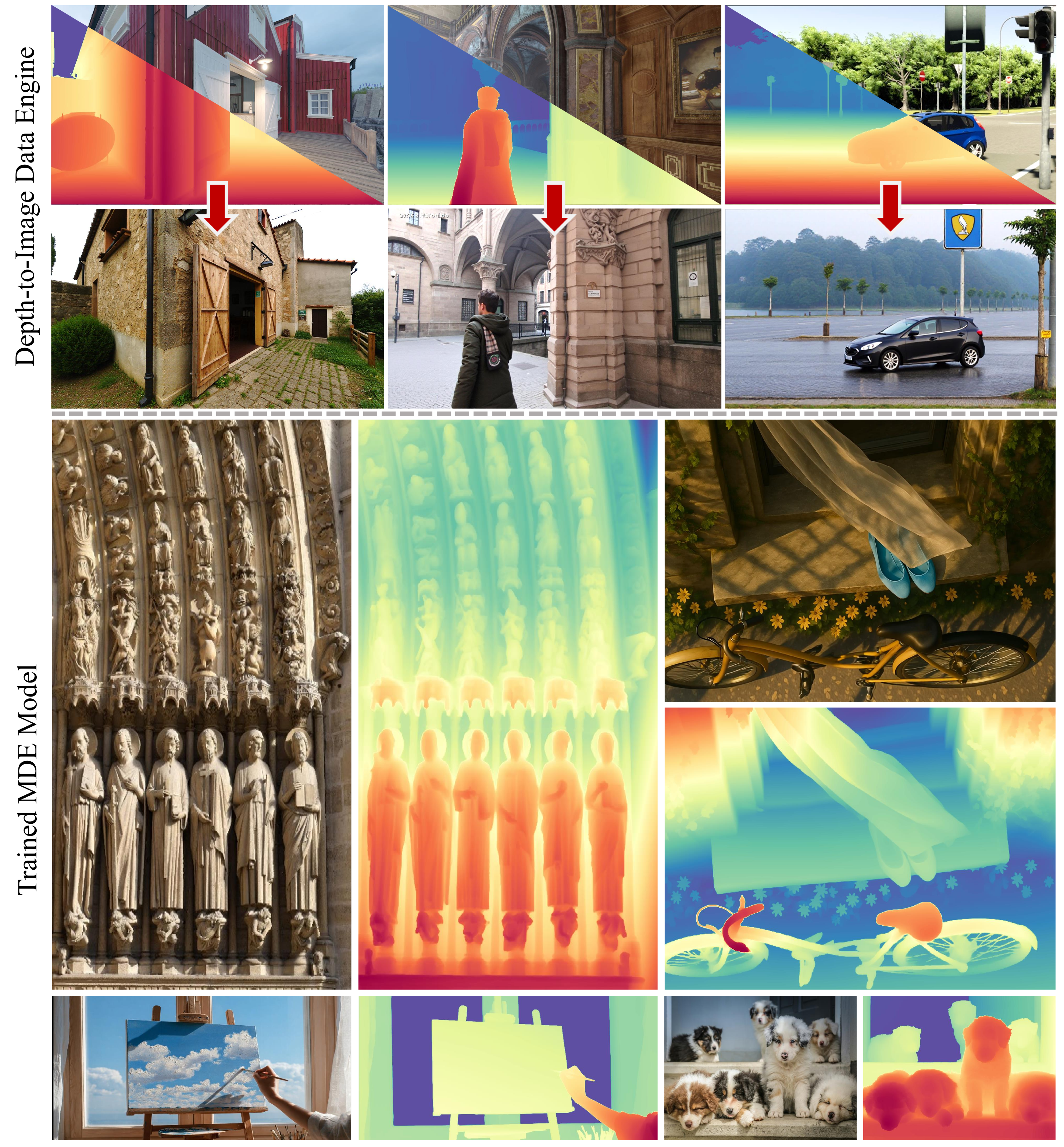}
    % \framebox[0.9\textwidth][c]{Placeholder for zero shot prediction visualization}
    \caption{We present \name, showcasing its RL-optimized Depth-to-Image (D2I) data generation engine which is used for generating realistic and geometrically accurate RGB images from source depth maps and Monocular Depth Estimation (MDE) model which after being trained on the massive high-quality data generated by the D2I engine, achieves superior depth prediction in complex scenes.}
    \label{fig:zero_shot_prediction}
\end{figure}
% \vspace{12pt}
\newpage
\begin{abstract}
Monocular Depth Estimation (MDE) is a foundational task for computer vision. Traditional methods are limited by data scarcity and quality, hindering their robustness. To overcome this, we propose \name, an RL-optimized depth-to-image (D2I) generation framework that synthesizes over 20M realistic and geometrically accurate RGB images, each intrinsically paired with its ground truth depth, from diverse source depth maps. Then we train our depth estimation model on this dataset, employing a hybrid supervision strategy that integrates teacher pseudo-labels with ground truth depth for comprehensive and robust training. This innovative data generation and training paradigm enables \name\ to achieve breakthroughs in scale and domain diversity, consistently outperforming existing state-of-the-art approaches quantitatively and in complex scene detail capture, thereby fostering general and robust depth features. Code and models are available at \text{\url{https://dingning-liu.github.io/bridge.github.io/}}.

\end{abstract}

\input{Chapter/1.intro}
\input{Chapter/2.related_works}
\input{Chapter/3.method}

\input{Chapter/4.exp}

\input{Chapter/5.conclusion}

\bibliography{main}
\bibliographystyle{main}

\newpage
\appendix
% \section{Appendix}

\input{Chapter/appendix}
% You may include other additional sections here.

\end{document}

%% file: math_commands.tex
\usepackage{amsmath,amsfonts,bm}

\def\eqref#1{equation~\ref{#1}}
\def\1{\bm{1}}

\DeclareMathAlphabet{\mathsfit}{\encodingdefault}{\sfdefault}{m}{sl}
\SetMathAlphabet{\mathsfit}{bold}{\encodingdefault}{\sfdefault}{bx}{n}

%% file: Chapter/1.intro.tex
% \newpage
\section{Introduction}
Monocular Depth Estimation (MDE) stands as a cornerstone in the field of 3D computer vision, providing essential geometric perception for various critical applications such as 3D reconstruction~\citep{mildenhall2021nerf,kerbl20233d,charatan2024pixelsplat,ye2024gaustudio}, autonomous driving~\citep{wang2019pseudo}, robotics~\citep{wofk2019fastdepth}, and AR/VR~\citep{rasla2022relative}. 
The field has witnessed notable progress in recent years~\citep{atapour2018real,ranftl2020towards,zoedepth,hu2024metric3d,rajpal2023high,yang2024depthvideo,ke2024repurposing,yang2024depth1,yang2024depth2,fu2024geowizard,wang2025moge}, however, training robust MDE models with excellent generalization remains challenging due to the scarcity of high-quality, precise ground truth depth annotations, insufficient detail and diversity in existing labels, and the underutilization of available depth data, creating a critical bottleneck for MDE model training.

Existing data acquisition and utilization methods can be categorized into three groups.
(1) \cite{zoedepth,hu2024metric3d,piccinelli2024unidepth} primarily rely on real-world data. While valuable, this approach is constrained by the sparsity of depth maps from sensors and the difficulty of accurately capturing and annotating transparent or reflective objects~\citep{costanzino2023learning}.

(2) \cite{atapour2018real,rajpal2023high,yang2024depthvideo,ke2024repurposing} are mainly based on synthetic data. Although synthetic data is theoretically precise, it may introduce significant domain gaps~\citep{ganin2015unsupervised} and geometric artifacts, degrading models' generalization capabilities.

(3) \cite{ranftl2020towards,yang2024depth1,yang2024depth2,fu2024geowizard,wang2025moge} integrate diverse data sources. MiDaS~\citep{ranftl2020towards} as the pioneer significantly enhances generalization by training on various mixed datasets; however, its data coverage and diversity remain limited. Building upon this, Depth Anything~\citep{yang2024depth1,yang2024depth2} advances SOTA by leveraging massive real-world images and a teacher model for pseudo-label generation, significantly improving generalization. However, its dependence on large real datasets and pseudo-label inaccuracies (despite high generation quality) bottleneck further performance and efficiency. These data-level limitations—diversity, scale, and pseudo-label fidelity—hinder truly universal and efficient MDE models.

To address the above challenges, we propose BRIDGE, an RL-optimized, large-scale Depth-to-Image (D2I) generation data engine. This engine leverages precise ground truth from synthetic depth datasets to generate massive, high-quality, and diverse RGB-D training data.
Data created by the data engine enable MDE models to achieve breakthroughs in both scale and domain diversity, thereby fostering general, robust depth features and excellent real-world performance.

Specifically, our core lies in an innovative data generation pipeline: we first train an RL-optimized Depth-to-Image (D2I) generation model on many synthetic datasets. This model can directly utilize diverse source depth maps (from existing synthetic depth datasets) to synthesize over 20 million diverse and information-rich RGB images. Based on this, we introduce a hybrid data supervision strategy. This strategy pairs generated RGB images with original high-precision ground truth depth, which is screened through similarity detection methods such as SSIM and gradient analysis. Simultaneously, it leverages a large number of pseudo-labels generated by a teacher model trained on synthetic data, and both are jointly used for training. This ensures that the generated RGB images receive reliable supervision during training while improving the overall quality and utilization efficiency of the labels.

Our experimental results consistently demonstrate \name's superior performance across various challenging benchmarks, including indoor, outdoor, and synthetic animation environments, surpassing existing state-of-the-art methods. Qualitatively, BRIDGE excels at capturing fine-grained details and maintaining robustness in complex scene structures. Furthermore, ablation studies confirm the effectiveness of the RL-D2I generated data and the hybrid supervision training method.

Our main contributions are as follows:
\begin{itemize}
    \item \textbf{An efficient data generation and utilization paradigm:} We adopt an RL-driven D2I paradigm, efficiently generating massive high-quality RGB-D data, which effectively alleviates data scarcity and quality issues.
    \item \textbf{Key practice of hybrid depth supervision training strategy:} We practice a hybrid depth supervision strategy, combining teacher model pseudo-labels with high-precision ground truth depth, to achieve a more challenging optimization objective and learn geometric structural knowledge from RGB data.
    \item \textbf{Superior performance and high training efficiency:} Our data generation and training strategy, using only approximately 20M data, surpasses SOTA models (e.g., Depth Anything V2 which uses 62M data), significantly improving training efficiency and demonstrating excellent detail capture capability and robustness.
\end{itemize}

%% file: Chapter/2.related_works.tex
\section{related works}
\subsection{Monocular Depth Estimation}
Monocular Depth Estimation (MDE), a foundational computer vision task, has seen significant progress with deep learning, moving beyond early methods limited by data homogeneity ~\citep{saxena2008make3d}. Deep learning approaches are broadly categorized into Discriminative Methods and Generative Methods. The former, starting with CNN-based pioneers like~\citep{krizhevsky2009learning,eigen2014depth,eigen2015predicting,chen2016single} and later enhanced by multi-dataset training in MiDaS~\citep{ranftl2020towards}, learns a direct mapping from image to depth~\citep{fu2018deep}. The adoption of the Transformer architecture~\citep{dosovitskiy2020image,ranftl2021vision,liu2021swin} and large-scale self-supervised learning, as in Depth Anything~\citep{yang2024depth1}, further improves generalization, while models like NDDdepth~\citep{shao2023nddepth} and Depth Pro ~\citep{bochkovskii2024depth} target high-precision boundary estimation. In contrast, generative methods utilize models like Diffusion Models~\citep{ho2020denoising,song2020denoising,nichol2021improved,ji2023ddp} to synthesize depth maps, with works such as LeReS~\citep{yin2021learning} and Metric3D v2 ~\citep{hu2024metric3d} focusing on accurate metric depth. The recent Depth Anything V2 ~\citep{yang2024depth2} combines large-scale synthetic images and pseudo-labeling to boost performance. Despite these advancements, a key challenge remains: achieving highly accurate and generalizable depth estimation in complex, unseen scenarios. To tackle this, we propose BRIDGE, a framework that generates a massive dataset of realistic and geometrically accurate RGB images to expand the effective training data.

\subsection{Self-supervised and Hybrid Supervision Learning}
To train a robust MDE model despite the scarcity of precisely labeled data, researchers have explored various weakly supervised and self-supervised approaches, such as utilizing unlabeled monocular or stereo video data through photometric consistency loss for training~\citep{godard2017unsupervised, kuznietsov2017semi,ren2020suw}. Other self-supervised approaches also include consistency training methods like FixMatch ~\citep{sohn2020fixmatch} and unsupervised data augmentation ~\citep{xie2020unsupervised}, as well as strategies like shrinking class space ~\citep{yang2023shrinking} to enhance certainty. Building on this, Pseudo-labeling supervision strategies~\citep{lee2013pseudo}, exemplified by Depth Anything V2~\citep{yang2024depth2}, significantly improve model generalization by employing a powerful teacher model to generate pseudo-labels for massive amounts of unlabeled images. However, despite the broad coverage provided by pseudo-labels, their inherent noise and inaccuracy—especially around boundaries and fine details—remain a bottleneck for further enhancing depth estimation performance. To leverage both the scale of pseudo-labels and the precision of ground truth annotations, we propose a hybrid supervision strategy. Unlike methods that solely rely on pseudo-labels, our strategy integrates high-precision ground truth depth from the original data via a similarity detection mechanism. By utilizing the teacher model's pseudo-labels for broad training and the similarity-guided ground truth for precision fine-tuning, our hybrid approach provides a more reliable and comprehensive supervision signal.

\subsection{Depth-to-Image Generation and Synthetic Data}
The performance of Monocular Depth Estimation (MDE) is directly linked to the scale and quality of training data, yet acquiring high-quality ground truth depth from real-world data is expensive and lacks sufficient diversity, leading researchers to increasingly utilize synthetic data. Traditional synthetic methods, which rely on 3D rendering engines, provide geometrically accurate annotations but often suffer from a pronounced domain gap~\citep{atapour2018real, rajpal2023high} and low generation efficiency. More recently, methods based on Generative Models have been explored, such as using Diffusion Models~\citep{ke2025marigold,gui2025depthfm} to generate depth maps from RGB images, but these primarily focus on depth prediction rather than the active generation of high-fidelity RGB-D data pairs for model training. To address this data bottleneck, Depth-to-Image (D2I) generation emerges as an active data augmentation strategy, synthesizing geometrically consistent RGB images from existing depth data. However, conventional D2I models often struggle with precise geometric alignment, leading to artifacts. Our BRIDGE framework alleviates this by employing a Reinforcement Learning (RL) optimization mechanism in its D2I engine, ensuring both visual realism and geometric consistency for the massive generated dataset.

%% file: Chapter/3.method.tex
\afterpage{\begin{figure}[!ht]
    \centering
\includegraphics[width=\linewidth]{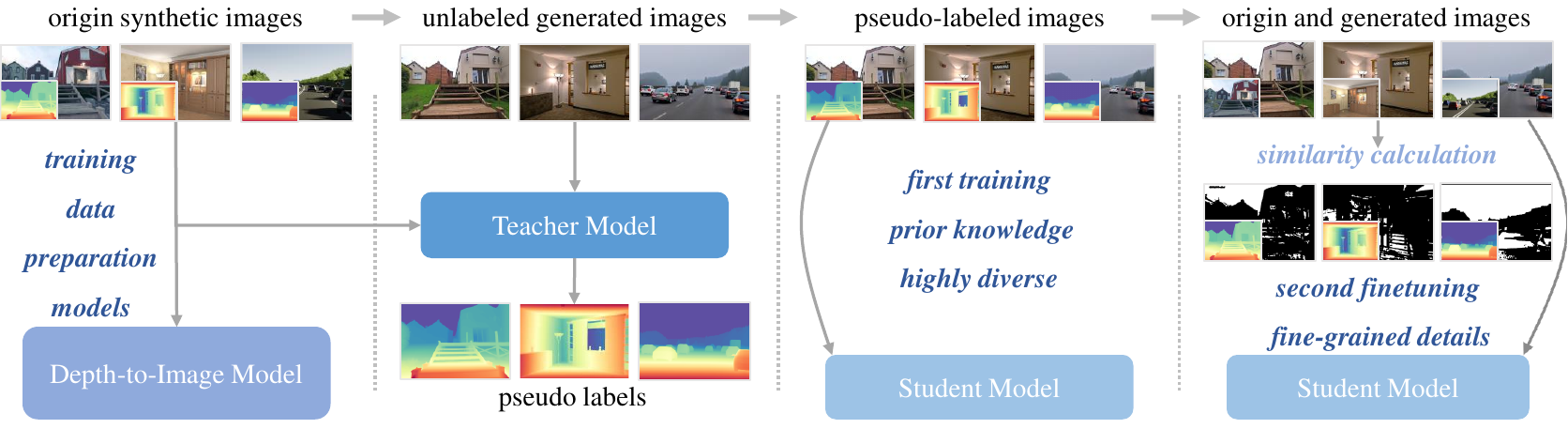}
    % \framebox[0.9\textwidth][c]{Placeholder for zero shot prediction visualization}
    \caption{\textbf{\name\ Pipeline.} First training a depth-to-image model to synthesize millions of realistic RGB images with precise ground truth depths and a teacher model for pseudo labeling. Student model is then trained on this extensive synthetic dataset. Finally, it's fine-tuned using mask-based refinement with original ground truth depth for robust generalization and detailed depth capture.}
    \label{fig:pipeline_overview}
\end{figure}
}
\section{Method}
To address data scarcity and label quality in monocular depth estimation, we propose a comprehensive three-stage pipeline, as illustrated in Figure~\ref{fig:pipeline_overview}.
First, we train a powerful giant teacher model on large-scale synthetic data, and train a depth-to-image model using reinforcement learning.
Second, we generate millions of visually realistic RGB images with accurate geometric structures, highly consistent with source depth maps, and their corresponding pseudo-labels, using the model we train in the previous stage (\Cref{sec:d2i}). Then we use a simple similarity algorithm method to calculate the original images and their corresponding generated images to create a mask for similar regions, and use the high-precision mask area and the original gt depth map for fine-tuning (\Cref{sec:pseudo_label}). Finally, we train the MDE model on the large-scale generated data and their pseudo-labels  (\Cref{sec:mde}).

\subsection{Reinforcement Learning-Optimized Depth-to-Image Generation}
\label{sec:d2i}
To acquire large-scale, high-quality RGB-depth image pairs, we develop and train a reinforcement learning (RL)-optimized Depth-to-Image (D2I) generation model. This model generates visually realistic RGB images from source depth maps (from existing synthetic datasets like Hypersim~\citep{roberts2021hypersim}, TartanAir~\citep{wang2020tartanair}) while precisely preserving their geometric structure. This generation process results in a large-scale dataset comprising approximately 20M images.

    \noindent \textbf{Objectives and Advantages:} Leveraging the reinforcement learning paradigm, we ensure generated images are not only visually realistic but also geometrically accurate and consistent. This enables synthesizing diverse RGB images from a single depth map while preserving its true geometric structure. This approach mitigates geometric artifacts or structural distortions common in traditional D2I models and expands the diversity of effective training data.

\begin{figure}[!ht]
    \centering
\includegraphics[width=\linewidth]{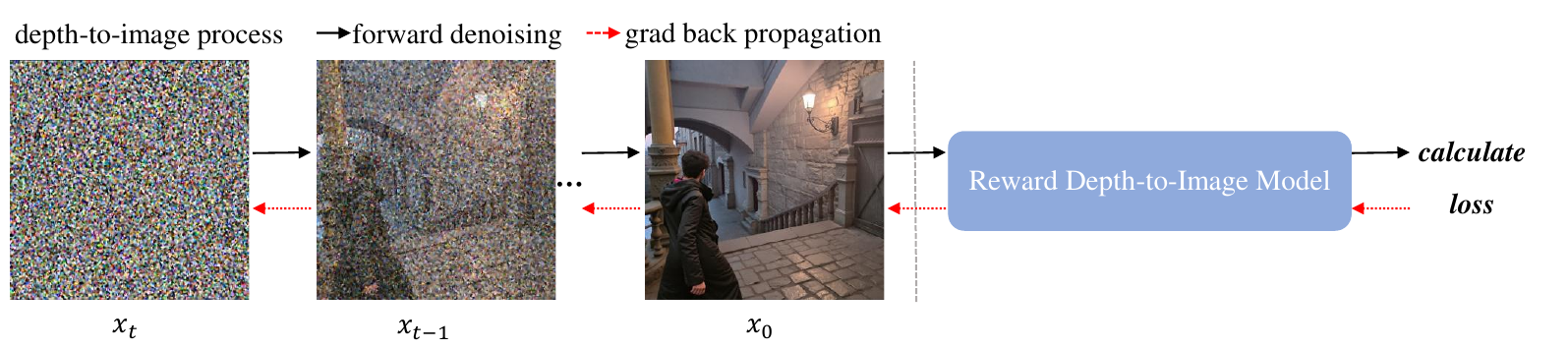}
    % \framebox[0.9\textwidth][c]{Placeholder for zero shot prediction visualization}
    \caption{\textbf{Reward Model Process.} Our D2I model is trained via reward-gradient-driven direct optimization , avoiding complex proxy objective functions and improving training memory efficiency.}

    \label{fig:rl_process}
\end{figure}

    \noindent \textbf{Training Strategy:} As shown in \Cref{fig:rl_process}, our D2I model, inspired by VADER~\citep{prabhudesai2024video}, is trained via reward-gradient-driven direct optimization to efficiently and accurately generate high-fidelity images. This direct usage of reward gradients guides the generation process towards desired visual quality, thereby avoiding complex proxy objective functions and improving training memory efficiency. The overall optimization objective minimizes the following loss:

    % \[ L_{\text{total}}(\theta) = \lambda_{\text{depth}} \cdot L_{\text{DMSE}}(\mathcal{M}(D_{\text{source}}, \theta), D_{\text{source}}) - \lambda_{\text{aesthetic}} \cdot R_{\text{aesthetic}}(\mathcal{M}(D_{\text{source}}, \theta)) \]
    \begin{equation}
        L_{\text{total}}(\theta) = \lambda_{\text{depth}} \cdot L_{\text{D}}(\mathcal{M}(D_{\text{source}}, \theta), D_{\text{source}}) - \lambda_{\text{aesthetic}} \cdot R_{\text{aesthetic}}(\mathcal{M}(D_{\text{source}}, \theta)),
    \end{equation}
    where $D_{\text{source}}$ represents the input source depth map, $\mathcal{M}(D_{\text{source}}, \theta)$ denotes the RGB image generated by the D2I model $\mathcal{M}$ with parameters $\theta$ based on $D_{\text{source}}$, $L_{\text{D}}(D_{\text{gen}}, D_{\text{source}})$ is a cosine similarity loss. Its corresponding loss (defined as $1 - \text{cosine similarity reward}$) aims to minimize the difference between the depth map $D_{\text{gen}}$ inverted from the generated RGB image and the source depth map $D_{\text{source}}$. This ensures the D2I model precisely learns and maintains the scene's geometric structural information. $R_{\text{aesthetic}}(\mathcal{M}(D_{\text{source}}, \theta))$ is the aesthetic reward function, quantifying the visual quality and aesthetic appeal of the generated image. It is formulated as:
    % \[ R_{\text{aesthetic}}(I) = g_{\text{MLP}}\left(\frac{f_{\text{CLIP}}(I)}{\|f_{\text{CLIP}}(I)\|_2}\right) \]
    \begin{equation}
        R_{\text{aesthetic}}(I) = g_{\text{MLP}}\Biggl(\frac{f_{\text{CLIP}}(I)}{\|f_{\text{CLIP}}(I)\|_2}\Biggr),
    \end{equation}
    
    where $f_{\text{CLIP}}(I)$ represents the high-dimensional feature embedding extracted from the input image $I$ by a pre-trained CLIP image encoder. $\|\cdot\|_2$ denotes the L2 norm, normalizing the extracted features to unit length to prevent their magnitude from unduly influencing subsequent scoring and to align with the MLP's training process. Finally, $g_{\text{MLP}}(\cdot)$ is a specialized MLP head that receives the normalized CLIP features as input and maps them to a scalar aesthetic score $S_{\text{aesthetic}} \in \mathbb{R}$. During D2I model optimization, we maximize the aesthetic quality of generated images by minimizing the aesthetic loss term $L_{\text{aesthetic}}(I) = -R_{\text{aesthetic}}(I)$, thereby incentivizing the model to produce images with higher aesthetic scores.

    The model is trained using gradient descent, with the parameter update rule:
    % \[\theta \leftarrow \theta - \eta \cdot \nabla_\theta L_{\text{total}}(\theta)\]
    \begin{equation}
        \theta \leftarrow \theta - \eta \cdot \nabla_\theta L_{\text{total}}(\theta),
    \end{equation}
    where $\eta$ is the learning rate. The core challenge is computing the gradient of the loss function with respect to model parameters $\theta$, $\nabla_\theta L_{\text{total}}(\theta)$. Since $x_0$ is the final output of the diffusion model's multi-step denoising process, and model parameters $\theta$ are involved in each denoising prediction $\epsilon_\theta(x_t, t, D_{\text{source}})$, the gradient must be backpropagated through the entire reverse diffusion process. Specifically, the gradient of the loss with respect to parameters can be expressed as:

    % \[\nabla_\theta L_{\text{total}}(\theta) = \sum_{t=0}^{T} \frac{\partial L_{\text{total}}(x_0)}{\partial x_t} \cdot \frac{\partial x_t}{\partial \theta}\]
    \begin{equation}
        \nabla_\theta L_{\text{total}}(\theta) = \sum_{t=0}^{T} \frac{\partial L_{\text{total}}(x_0)}{\partial x_t} \cdot \frac{\partial x_t}{\partial \theta},
    \end{equation}
    where $\frac{\partial L_{\text{total}}(x_0)}{\partial x_t}$ represents the sensitivity of the loss to the intermediate diffusion step $x_t$, and $\frac{\partial x_t}{\partial \theta}$ denotes the influence of model parameters on the intermediate state. This gradient propagation efficiently guides the diffusion model to generate high-quality RGB images while precisely capturing and maintaining the geometric structure of the input depth map, significantly enhancing training sample and computational efficiency.

\subsection{Depth Pseudo-label Generation and Multi-source Depth Fusion Strategy}
\label{sec:pseudo_label}
After generating millions of high-fidelity RGB images, we construct their corresponding depth labels using a multi-strategy fusion approach. These are then combined with the original synthetic data to form the final hybrid training set.

    \noindent \textbf{Teacher Model-Based Initial Pseudo-Label Generation:} We leverage a powerful teacher model trained on 1M synthetic data to infer initial relative depth pseudo-labels for all 20 million generated RGB images. These pseudo-labels provide valuable, geometrically consistent supervision, crucial for bridging the domain gap.
    
    \noindent \textbf{Similarity-Guided Ground Truth Depth Utilization:} To achieve more precise and reliable depth labels and maximize the use of all available depth information (including original high-precision ground truth depth), we introduce a similarity-based method. This addresses pixel-level inaccuracies or detail loss from relying solely on teacher-generated pseudo-labels. Specifically, we construct a fusion mask for each generated RGB image $I_{\text{gen}}$ by evaluating its similarity to the original synthetic RGB image $I_{\text{orig}}$. This mask generation involves two main steps:
        \begin{itemize}
            \item \textbf{Feature-based Registration for Structural Similarity:} First, we geometrically register the generated RGB image ($I_{\text{gen}}$) with its original synthetic counterpart ($I_{\text{orig}}$) using ORB feature detection and matching~\citep{karami2017image}. This step compensates for minor shifts or deformations during image generation, aligning them at the pixel level. After registration, we compute the Structural Similarity Index Measure (SSIM) map~\citep{nilsson2020understanding} between the aligned images. Regions with an SSIM value above a predefined threshold are identified as high-similarity areas, forming the first binary mask.
            \item \textbf{Direct SSIM-based Pixel-level Similarity:} Second, we directly compute the SSIM map between the size-adjusted generated RGB image ($I_{\text{gen}}$) and the original synthetic RGB image ($I_{\text{orig}}$). This evaluates their pixel-level similarity without explicit geometric registration. Regions where the SSIM value exceeds the same threshold form the second binary mask.
        \end{itemize}
    The final fusion mask is generated by applying a logical OR operation to these two binary masks. This ensures that any high-similarity region identified by either method is included. To enhance mask robustness and reduce noise, we further apply morphological opening and closing operations to smooth boundaries and fill small gaps.
    
    Upon obtaining the final fusion mask, we integrate this refined depth supervision into a two-stage training process. Initially, the \name\ model undergoes pre-training on a large-scale dataset, primarily using the teacher-generated pseudo-labels. Following this, a fine-tuning stage is performed. In this fine-tuning stage, for regions covered by the fusion mask, we directly utilize the corresponding high-precision ground truth depth ($D_{\text{gt}}$) from the original synthetic data as training labels. This strategy ensures the model first learns broad geometric consistency from vast pseudo-labeled data and then refines its accuracy and detail with the most accurate supervision in high-precision, geometrically consistent areas, thereby significantly improving depth estimation accuracy and detail.

    \noindent \textbf{Final Hybrid Dataset Composition:} The composite dataset for training \name\ comprises two main components. The first consists of large-scale generated RGB images: approximately 20 million high-fidelity RGB images produced by the RL-optimized D2I process, each paired with its teacher-model-derived depth pseudo-label. The second involves original synthetic data depth labels: we directly integrate depth labels from multiple original synthetic datasets, crucially including the high-precision ground truth depth filtered by our similarity-guided strategy.
    
\subsection{Training Monocular Depth Estimation Model}
\label{sec:mde}
% Our research primarily focuses on the effectiveness of the data generation strategy.
After designing the data generation strategy, we train the Monocular Depth Estimation (MDE) model on the generated data.
For model architecture, we adopt the established pretrained DINOv2-Giant encoder~\citep{oquab2023dinov2} and DPT~\citep{ranftl2021vision} head combination from Depth Anything~\citep{yang2024depth1}. To optimize our model, we utilize two loss terms, both of which are based on the approach proposed in MiDaS\citep{ranftl2020towards} because they are well-suited for  affine-invariant depth. The first is the scale- and shift-invariant loss ($L_{ssi}$), which ensures the model's robustness to variations in the scale and offset of depth values. The second is the gradient matching loss ($L_{gm}$), which helps the model capture fine-grained scene structure and details by enforcing similarity between the gradients of the predicted and ground-truth depth maps. We set the $L_{ssi}~\text{and}~L_{gm}$ as 1:4 and we ignore its top n largest loss regions during training, where n is set as 10\% because the regions are usually considered as potentially noisy pseudo labels. To enable zero-shot metric depth estimation for arbitrary ``in-the-wild'' images, we further introduce a specialized scale head. This head autonomously predicts the image's metric scale, significantly enhancing the model's accuracy and generalization in real-world scenarios. We first fine-tune the metric depth model, then decouple and train the Scale Head separately. This improves training efficiency and provides high-quality, artifact-free, precise metric depth for downstream tasks like novel view synthesis, further enhancing the model's utility.

%% file: Chapter/4.exp.tex
\afterpage{
\begin{table}[ht]
    \caption{\textbf{Quantitative Comparison with SOTA Methods on Zero-Shot Relative Depth Estimation.} These benchmarks aim to comprehensively measure model generalization ability and accuracy in diverse environments. While these standard benchmarks provide a direct quantitative comparison, some specific strengths of a model, such as details, complex layouts, may not be fully reflected by these benchmarks.The underline represents the best performance of SOTA models on this dataset.}
    \centering
    \resizebox{\linewidth}{!}{
    \setlength{\tabcolsep}{0.5mm} 
    \renewcommand{\arraystretch}{1.5}

    \begin{tabular}{l@{\hskip 15pt}l c@{\hskip 2pt}c c@{\hskip 2pt}c c@{\hskip 2pt}c c@{\hskip 2pt}c c@{\hskip 2pt}c c}
         \toprule
         \multirow{2}{*}{ } & \multirow{2}{*}[-14pt]{Method} & \multicolumn{2}{c}{KITTI} & \multicolumn{2}{c}{NYUv2} & \multicolumn{2}{c}{ScanNet} & \multicolumn{2}{c}{ETH3D} & \multicolumn{2}{c}{Sintel} & DA2K \\
         \cmidrule(lr){3-4} \cmidrule(lr){5-6} \cmidrule(lr){7-8} \cmidrule(lr){9-10} \cmidrule(lr){11-12} \cmidrule(lr){13-13} % 
         & & $\delta_1\uparrow$ & AbsRel$\downarrow$ & $\delta_1\uparrow$ & AbsRel$\downarrow$ & $\delta_1\uparrow$ & AbsRel$\downarrow$ & $\delta_1\uparrow$ & AbsRel$\downarrow$ & $\delta_1\uparrow$ & AbsRel$\downarrow$ & Acc(\%) \\
         \midrule 
         \multirow{3}{*}{\rotatebox[origin=c]{90}{\textbf{\textit{Generative}}}} & DepthFM~\citep{gui2025depthfm} & 0.932 & 0.085 & 0.956 & 0.065 & 0.947 & 0.068 & 0.958 & 0.069 & 0.700 & 0.577 & 85.8 \\
         & Marigold~\citep{ke2024repurposing} & 0.916 & 0.099 & 0.964 & 0.055 & 0.951 & 0.064 & 0.960 & 0.065 & 0.703 & 0.576 & 86.8 \\
         & GeoWizard~\citep{fu2024geowizard} & 0.921 & 0.097 & 0.966 & 0.052 & 0.953 & 0.061 & 0.961 & 0.064 & 0.705 & 0.574 & 88.1 \\
         \midrule
         \multirow{6}{*}{\rotatebox[origin=c]{90}{\textbf{\textit{Discriminative}}}} & LeReS~\citep{yin2021learning} & 0.784 & 0.149 & 0.916 & 0.090 & 0.917 & 0.091 & 0.777 & 0.171 & 0.590 & 0.785 & - \\
         & DPT~\citep{ranftl2021vision} & 0.881 & 0.111 & 0.919 & 0.091 & 0.932 & 0.084 & 0.929 & 0.115 & 0.605 & 0.700 & - \\
         & MiDaSv3.1~\citep{birkl2023midas} & 0.851 & 0.126 & 0.980 & 0.048 & 0.957 & 0.069 & 0.947 & 0.078 & 0.670 & 0.611 & - \\
         & Metric3Dv2~\citep{hu2024metric3d} & \underline{0.968} & 0.062 & 0.963 & 0.058 & 0.941 & 0.074 & 0.960 & 0.066 & 0.662 & 0.619 & - \\
         & Depth Anything V2~\citep{yang2024depth2} & 0.946 & 0.075 & 0.979 & 0.045 & 0.978 & 0.043 & 0.988 & 0.038 & 0.672 & 0.598 & 97.1 \\
        & Depth Pro~\citep{bochkovskii2024depth} & 0.498 & 0.375 & 0.580 & 0.245 & 0.667 & 0.207 & 0.687 & 0.201 & 0.476 & 0.883 & - \\

         & \textbf{Ours} & \textbf{0.938} & \textbf{0.081} & \textbf{0.982} & \textbf{0.041} & \textbf{0.981} & \textbf{0.033} & \textbf{0.991} & \textbf{0.029} & \textbf{0.719} & \textbf{0.513} & \textbf{97.3} \\
         \bottomrule 
    \end{tabular}
    } 
    \label{tab:relative_depth}
\end{table}
}

\section{Implementation Details}  
Our RL-D2I generative model is based on the architecture of Flux.1 Depth[dev]~\citep{flux2024} and fine-tuned using depth data from synthetic datasets, and the learning rate is 1e-5. The generator synthesizes approximately 20 million visually realistic and geometrically accurate RGB images from source depth maps by optimizing a loss function that includes depth loss (weight 0.9) and aesthetic reward (weight 0.1). To achieve high geometric accuracy and depth similarity, we ensure that the generated images meet stringent quality criteria: high-precision regions are identified by combining areas where: 1) ORB feature-based registered SSIM between the generated and original RGB images exceeds 0.85 (requiring at least 10 ORB matches), and 2) direct SSIM between them also exceeds 0.85. We only select mask samples where the valid region constitutes over 50\% of the pixels, and a 3x3 erosion operation is performed to filter out excessively small regions. Subsequently, a 518x518 crop is randomly extracted, centered on the largest valid region's bounding box. For each original depth map, we generate four RGB images simultaneously, and use different random seeds. The prompt is set to None during the generation process.

\section{Experiments}  
In this section, we will validate the effectiveness and advantages of the proposed method for monocular depth estimation through a series of experiments. We first introduce the datasets, evaluation metrics, and experimental settings. Then, we present the performance of our method on multiple benchmark datasets and conduct ablation experiments to analyze the contribution of each module to the overall performance.
\subsection{Evaluation Datasets}  
For monocular depth estimation, we conduct a series of experiments to evaluate the performance of our model on five widely used benchmarks. NYUv2~\citep{silberman2012indoor} and ScanNet~\citep{dai2017scannet} provide indoor RGB-D data captured by depth sensors. ETH3D~\citep{schops2017multi} includes both indoor and outdoor scenes, with depth data collected via laser scanners. KITTI~\citep{geiger2012we} consists of outdoor driving scenes captured with cameras and LiDAR sensors. For ScanNet, we extract 1400 test frames. Sintel~\citep{butler2012naturalistic} is a synthetic dataset derived from animated short films, from which we extract images of 600 training samples for evaluation.
\subsection{Evaluation Metrics}
All evaluations are carried out under the zero-shot configuration. Following previous works~\citep{yang2024depth2,he2025distill}, the evaluation of affine-invariant inverse depth prediction is performed by optimizing the scale and shift discrepancies between the estimated depth and the ground truth. For quantitative analysis, we utilize the absolute relative error (AbsRel) and $\delta$1 accuracy. AbsRel is calculated as $\frac{1}{N} \sum_{k=0}^{N-1} \frac{|\hat{x}_d - x_d|}{x_d}$, where $N$ represents the total pixel count. $\delta$1 accuracy measures the percentage of pixels where the maximum ratio between the predicted affine-invariant inverse depth and the inverse true depth falls below $1.25$.
\subsection{Zero-shot Depth Estimation}
Our model demonstrates exceptional zero-shot generalization in depth estimation across both indoor and outdoor datasets, highlighting its robustness, fine-grained detail, and consistent depth estimation for objects. 

\afterpage{\begin{figure}[!ht]
    \centering
    
    % Define column separation for compactness
    \setlength{\columnsep}{0.5mm}
    \setlength{\rowlabelcolwidth}{0.35cm}
    \setlength{\imagecolwidth}{\dimexpr (\textwidth - \rowlabelcolwidth - 5\columnsep) / 5 \relax} % 
    
    \makebox[\rowlabelcolwidth][l]{}%  <-- This reserves space for the row label
    \hspace{\columnsep}%     
    \makebox[\imagecolwidth]{\small\Centering RGB Image}%
    \hspace{\columnsep}%
    \makebox[\imagecolwidth]{\small\Centering Depth Anything V2}%
    \hspace{\columnsep}%
    \makebox[\imagecolwidth]{\small\Centering Depth Pro}%
    \hspace{\columnsep}%
    \makebox[\imagecolwidth]{\small\Centering Marigold}%
    \hspace{\columnsep}%
    \makebox[\imagecolwidth]{\small\Centering Ours}\\[1mm] 
    % First row: Kitti
    \raisebox{0.01\height}{\makebox[\rowlabelcolwidth][l]{\rotatebox{90}{\small KITTI}}}%
    \begin{subfigure}{\imagecolwidth}
    \includegraphics[width=\linewidth]{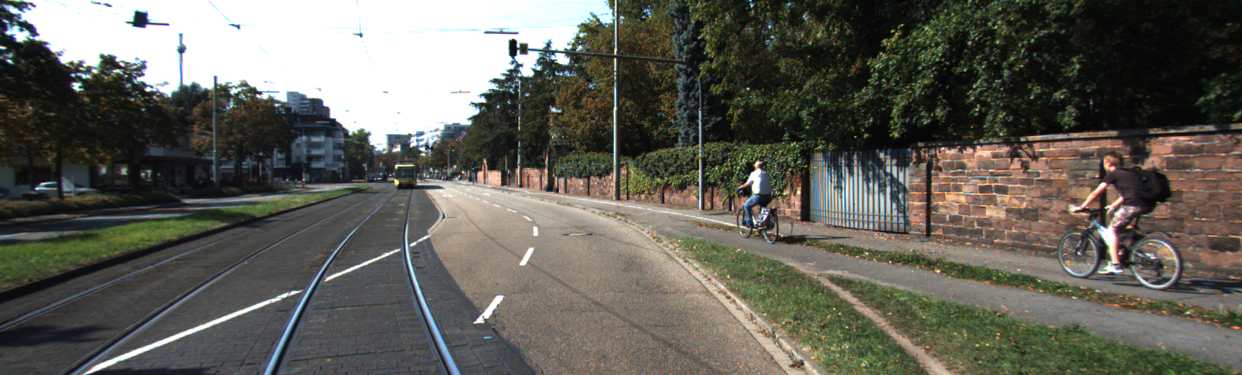}
    \end{subfigure}%
    \hspace{\columnsep}%
    \begin{subfigure}{\imagecolwidth}
    \includegraphics[width=\linewidth]{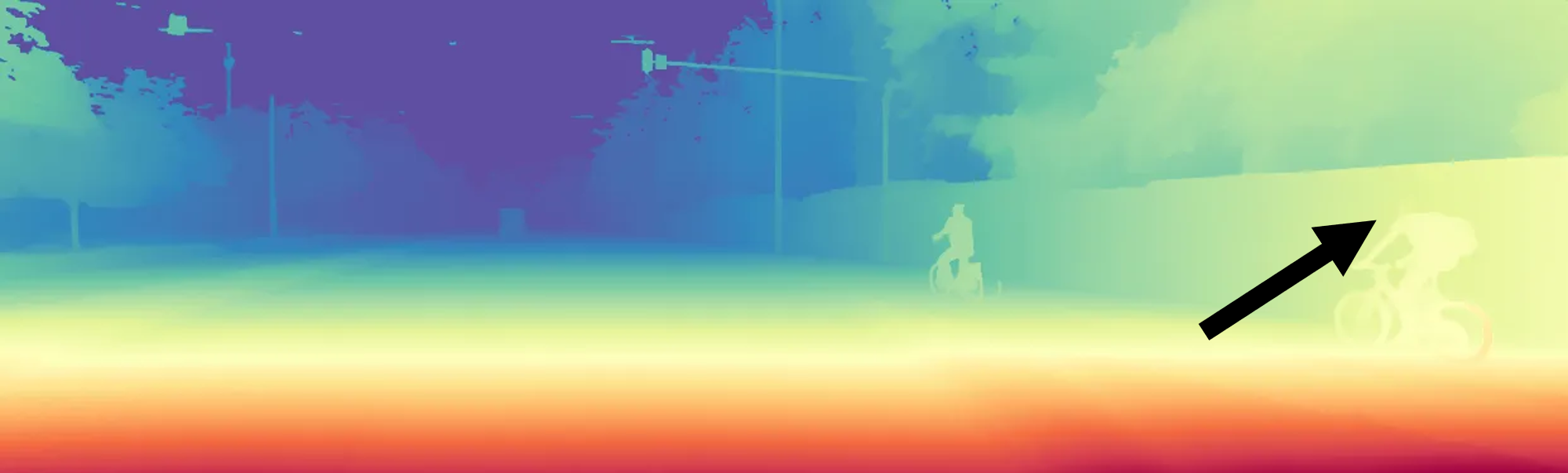}
    \end{subfigure}%
    \hspace{\columnsep}%
    \begin{subfigure}{\imagecolwidth}
    \includegraphics[width=\linewidth]{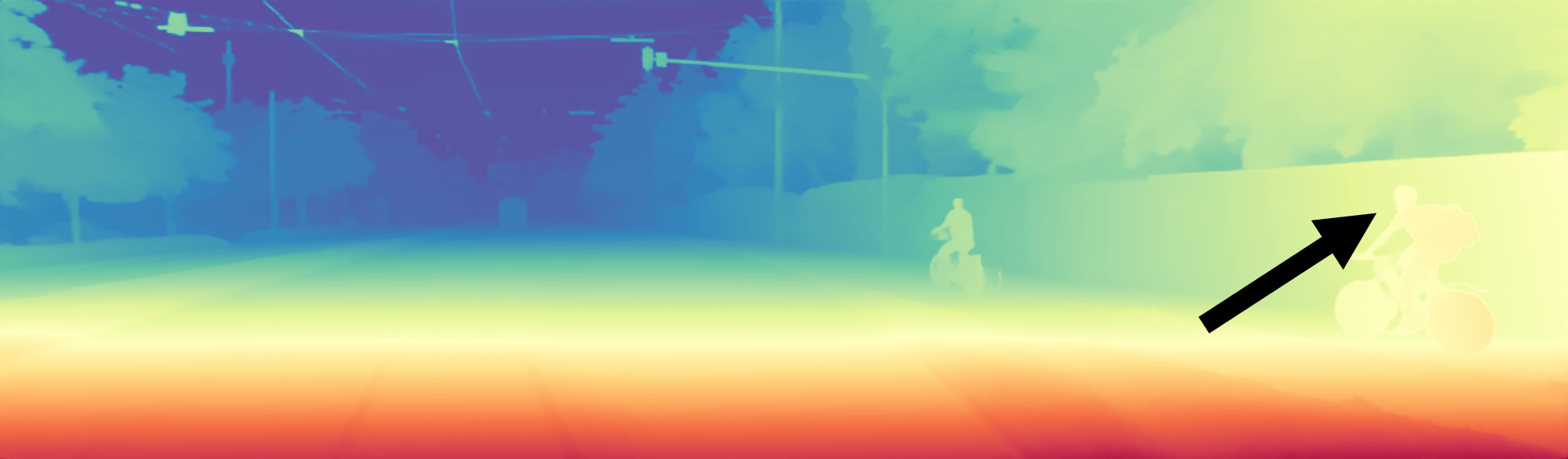} % Placeholder for Depth Pro image
    \end{subfigure}%
    \hspace{\columnsep}%
    \begin{subfigure}{\imagecolwidth}
    \includegraphics[width=\linewidth]{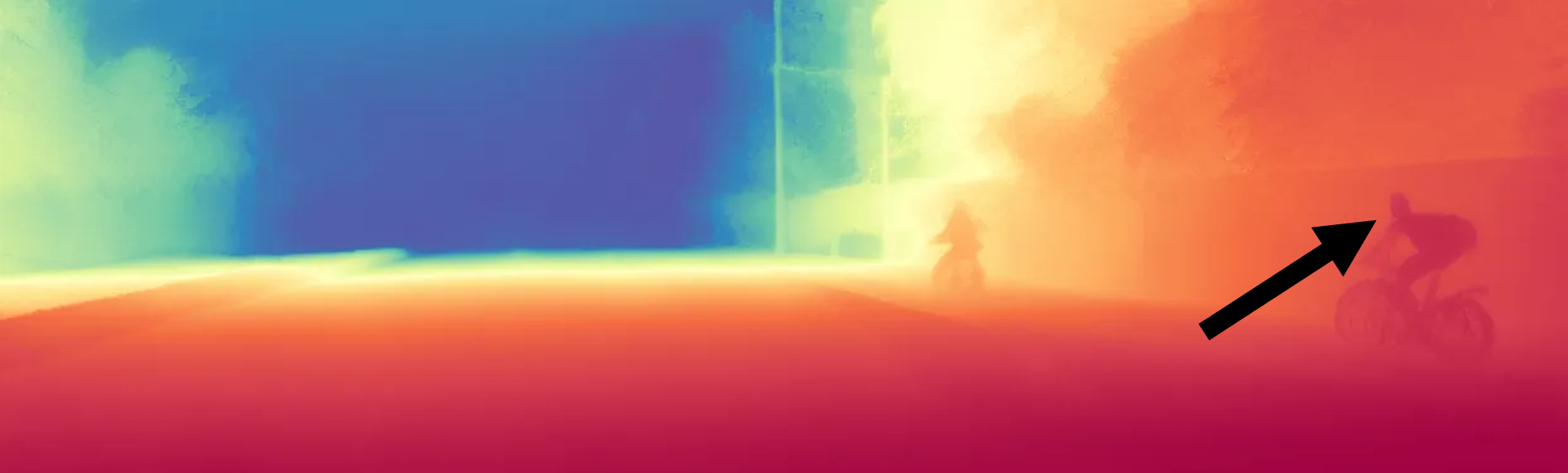}
    \end{subfigure}%
    \hspace{\columnsep}%
    \begin{subfigure}{\imagecolwidth}
    \includegraphics[width=\linewidth]{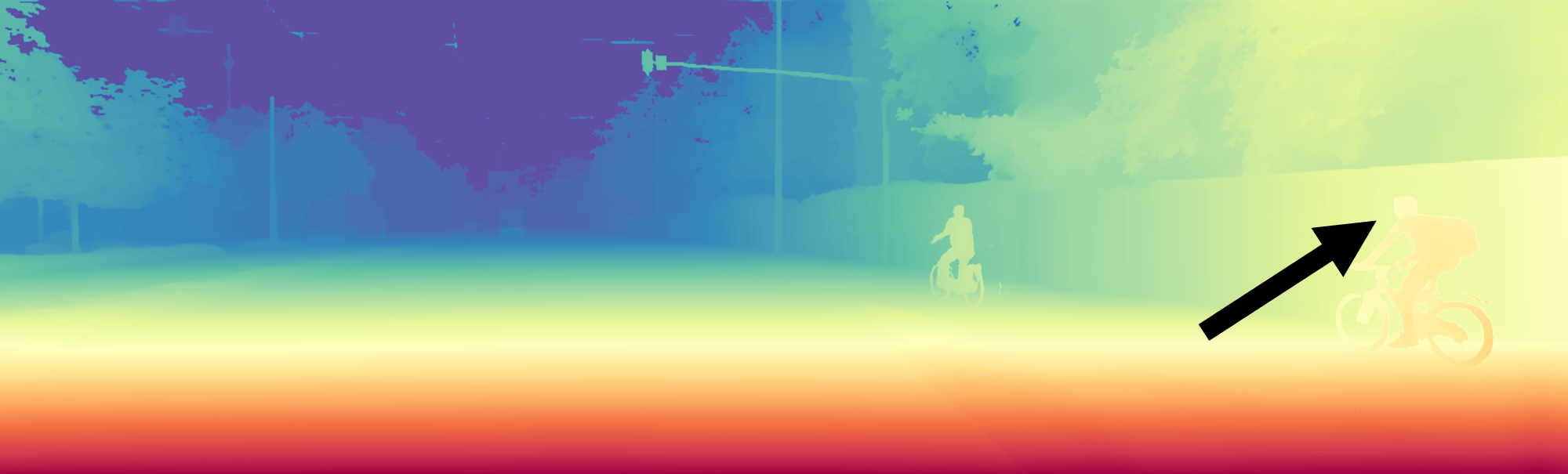}
    \end{subfigure}\\ % Reduced vertical space between rows
    
    % Second row: NYUv2
    \raisebox{0.4\height}{\makebox[\rowlabelcolwidth][l]{\rotatebox{90}{\small NYUv2}}}%
    \begin{subfigure}{\imagecolwidth}
    \includegraphics[width=\linewidth]{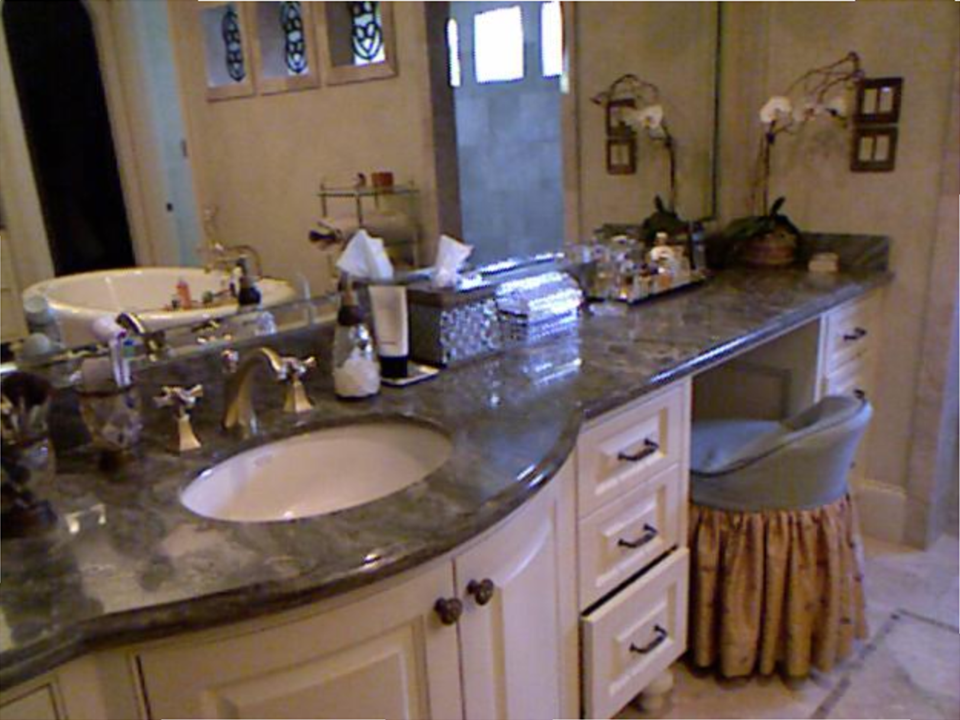}
    \end{subfigure}%
    \hspace{\columnsep}%
    \begin{subfigure}{\imagecolwidth}
    \includegraphics[width=\linewidth]{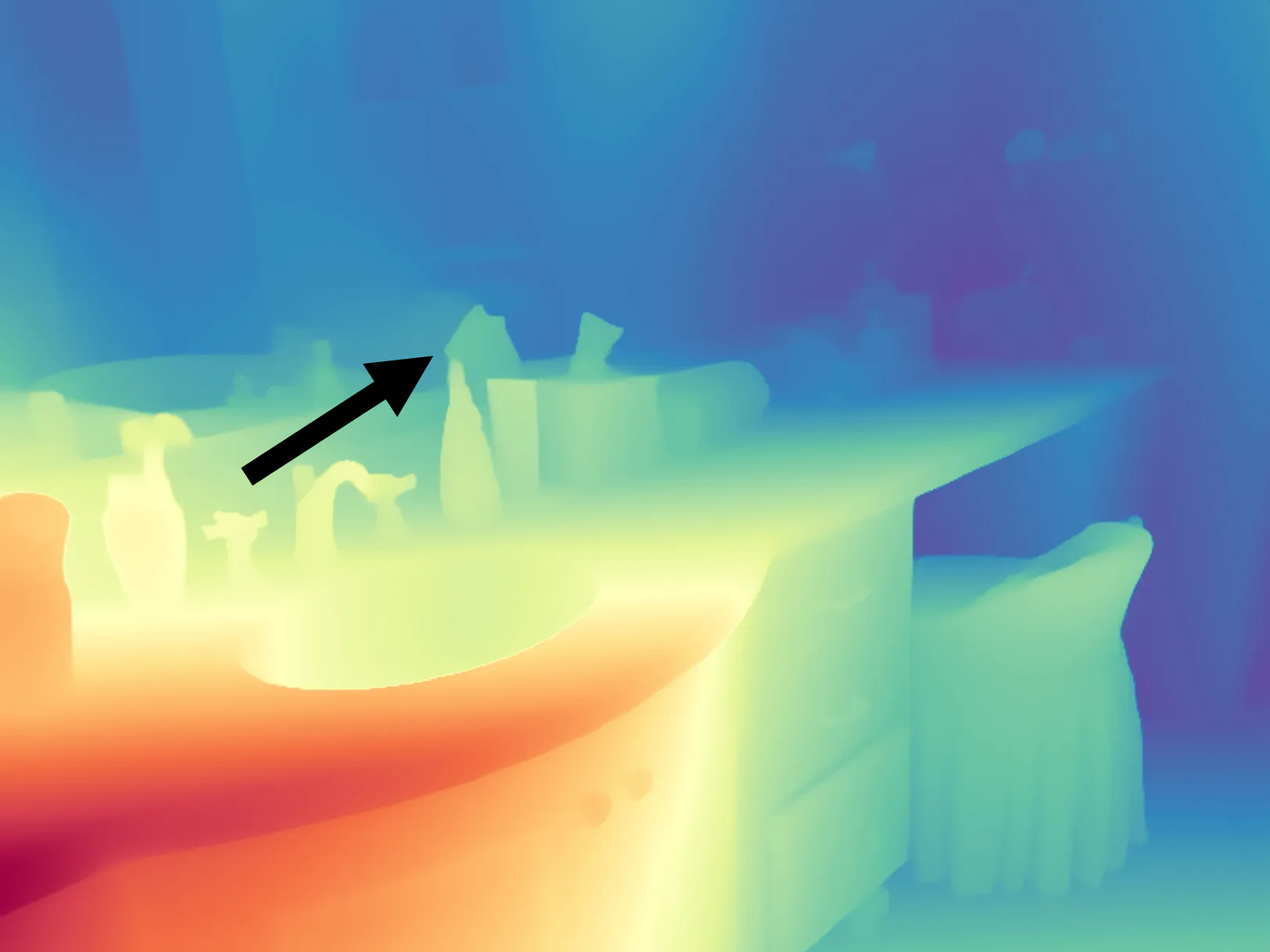}
    \end{subfigure}%
    \hspace{\columnsep}%
    \begin{subfigure}{\imagecolwidth}
    \includegraphics[width=\linewidth]{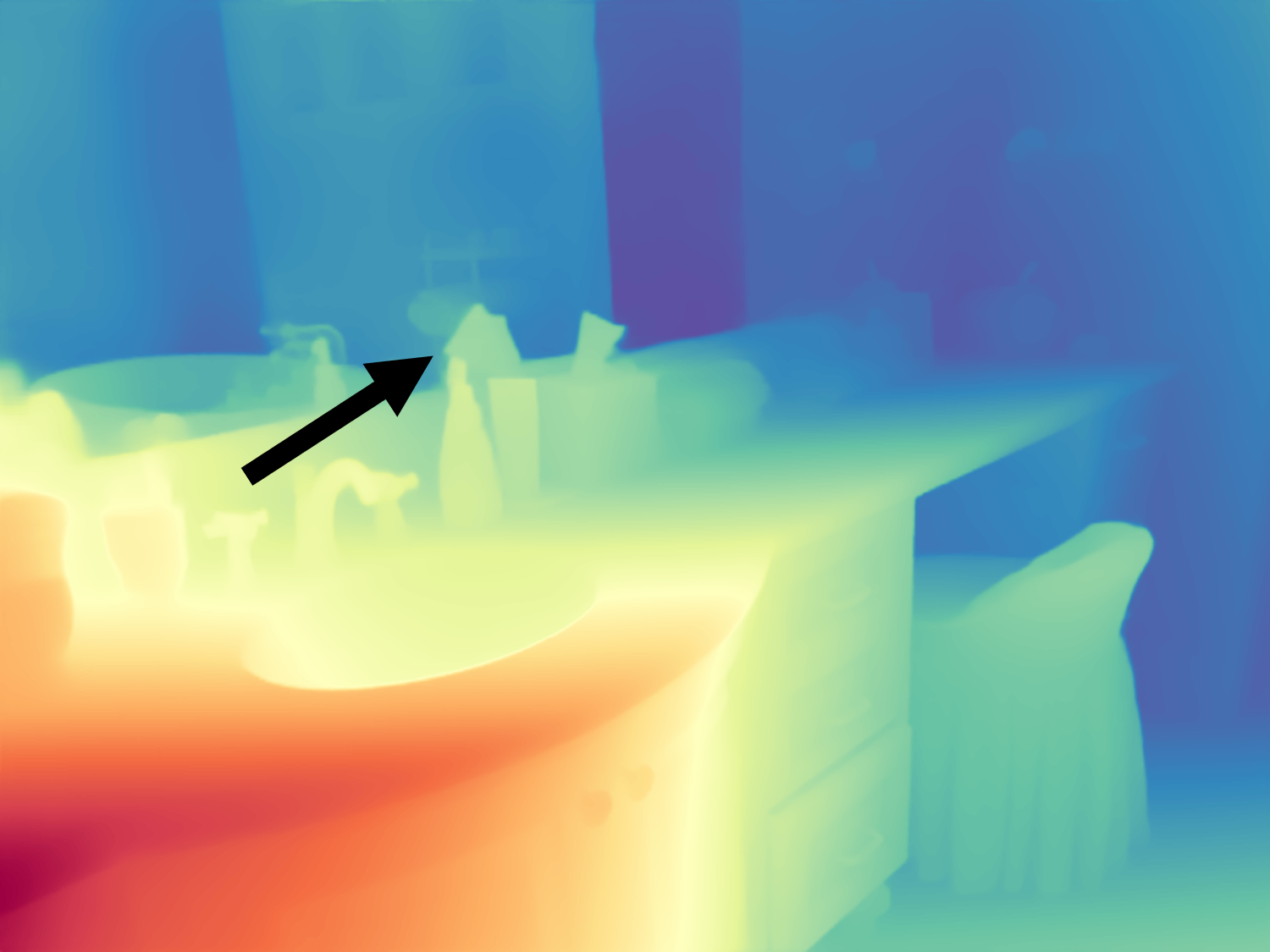} % Placeholder for Depth Pro image
    \end{subfigure}%
    \hspace{\columnsep}%
    \begin{subfigure}{\imagecolwidth}
    \includegraphics[width=\linewidth]{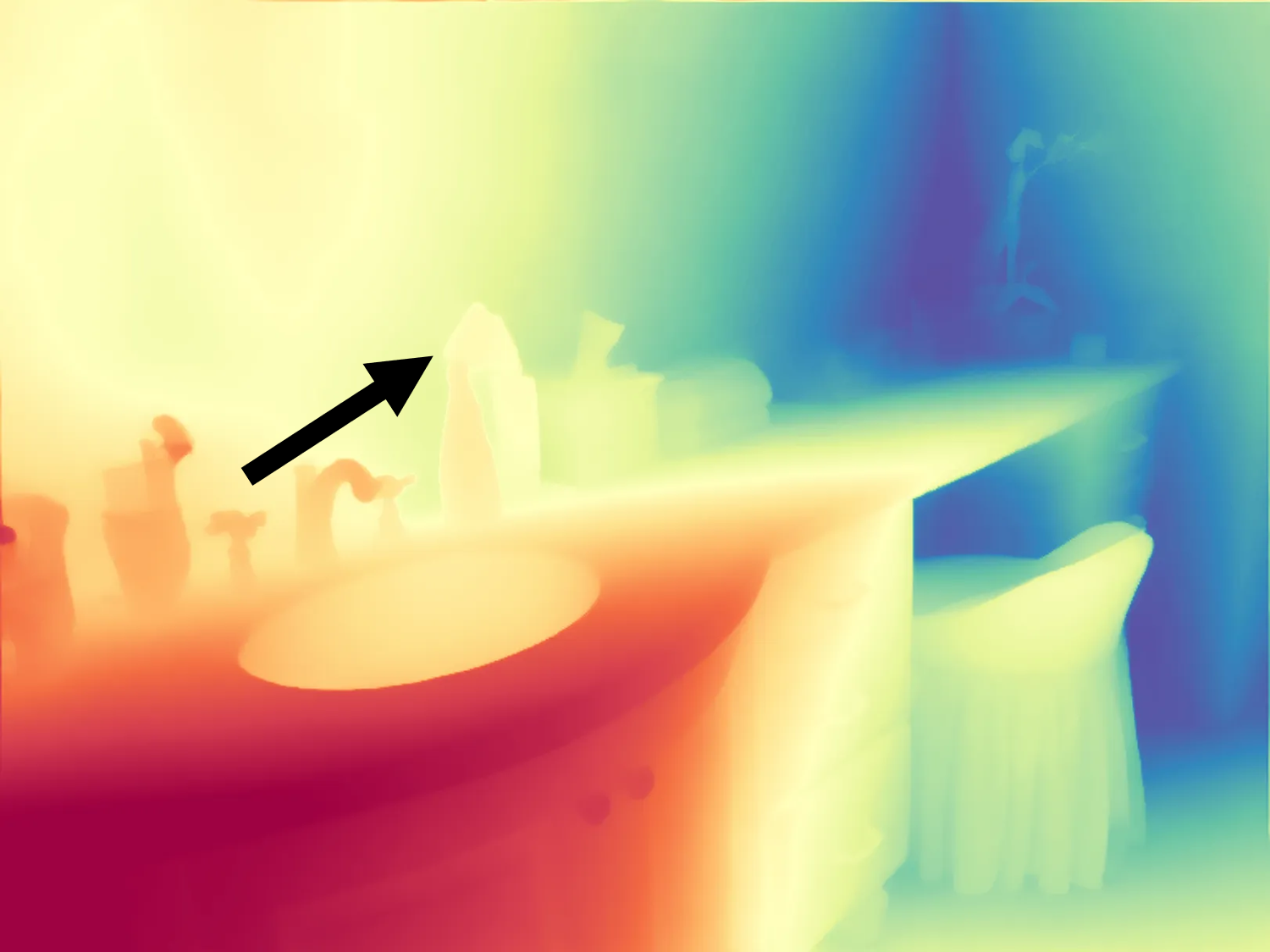}
    \end{subfigure}%
    \hspace{\columnsep}%
    \begin{subfigure}{\imagecolwidth}
    \includegraphics[width=\linewidth]{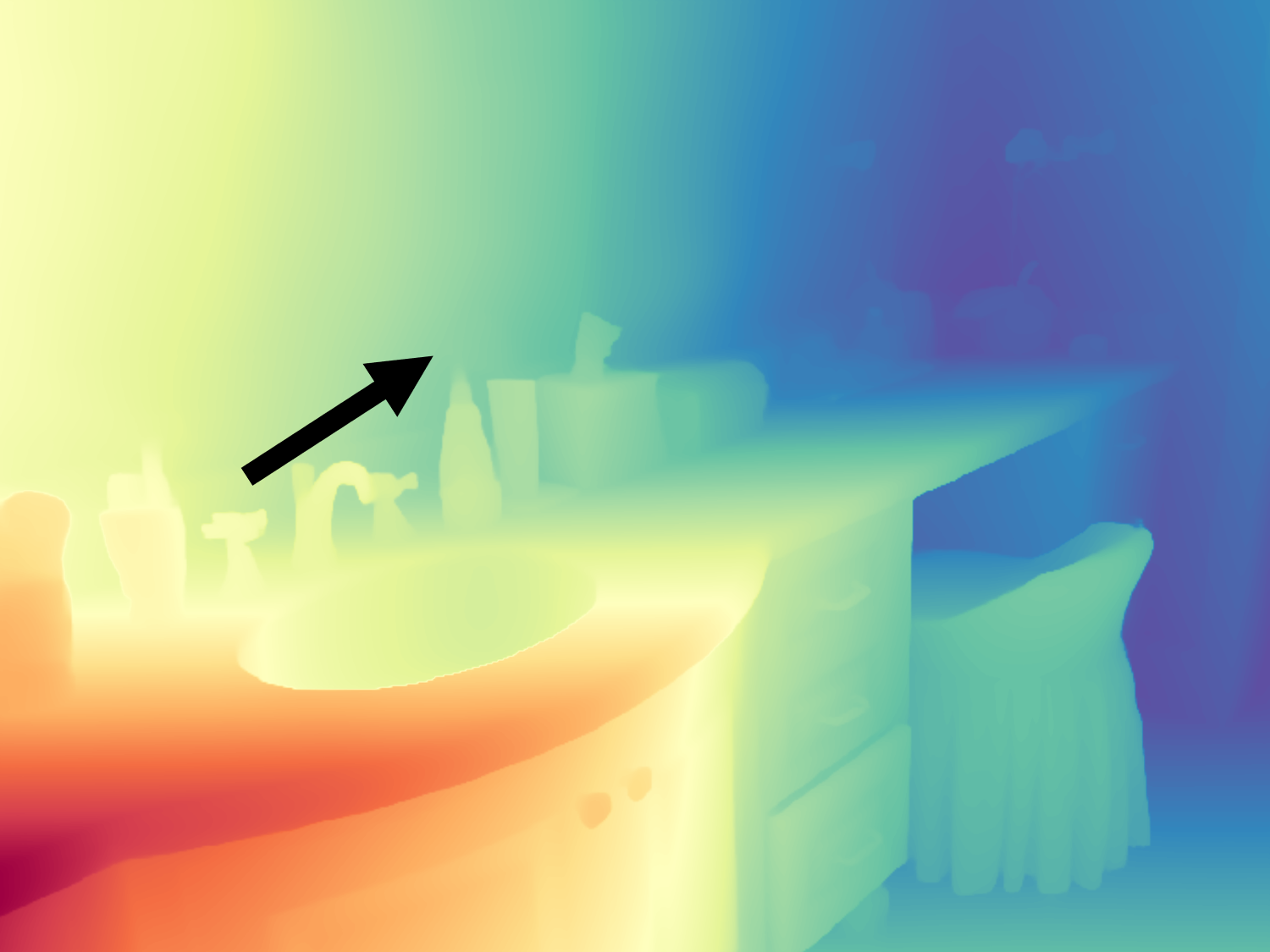}
    \end{subfigure}\\
    
    % Third row: ScanNet
    \raisebox{0.4\height}{\makebox[\rowlabelcolwidth][l]{\rotatebox{90}{\small ScanNet}}}%
    \begin{subfigure}{\imagecolwidth}
    \includegraphics[width=\linewidth]{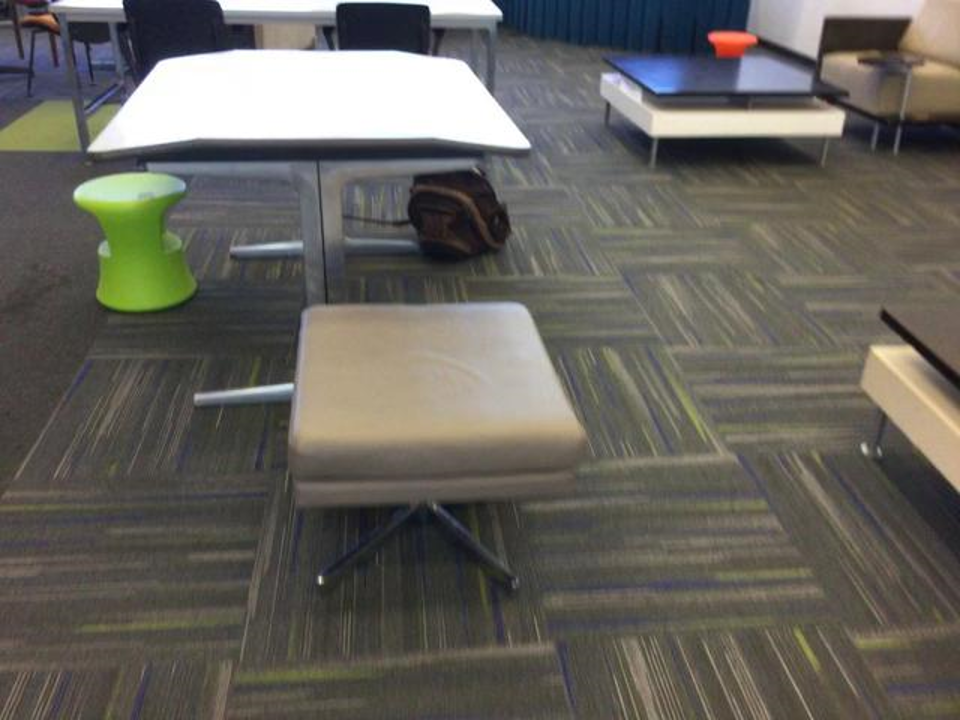}
    \end{subfigure}%
    \hspace{\columnsep}%
    \begin{subfigure}{\imagecolwidth}
    \includegraphics[width=\linewidth]{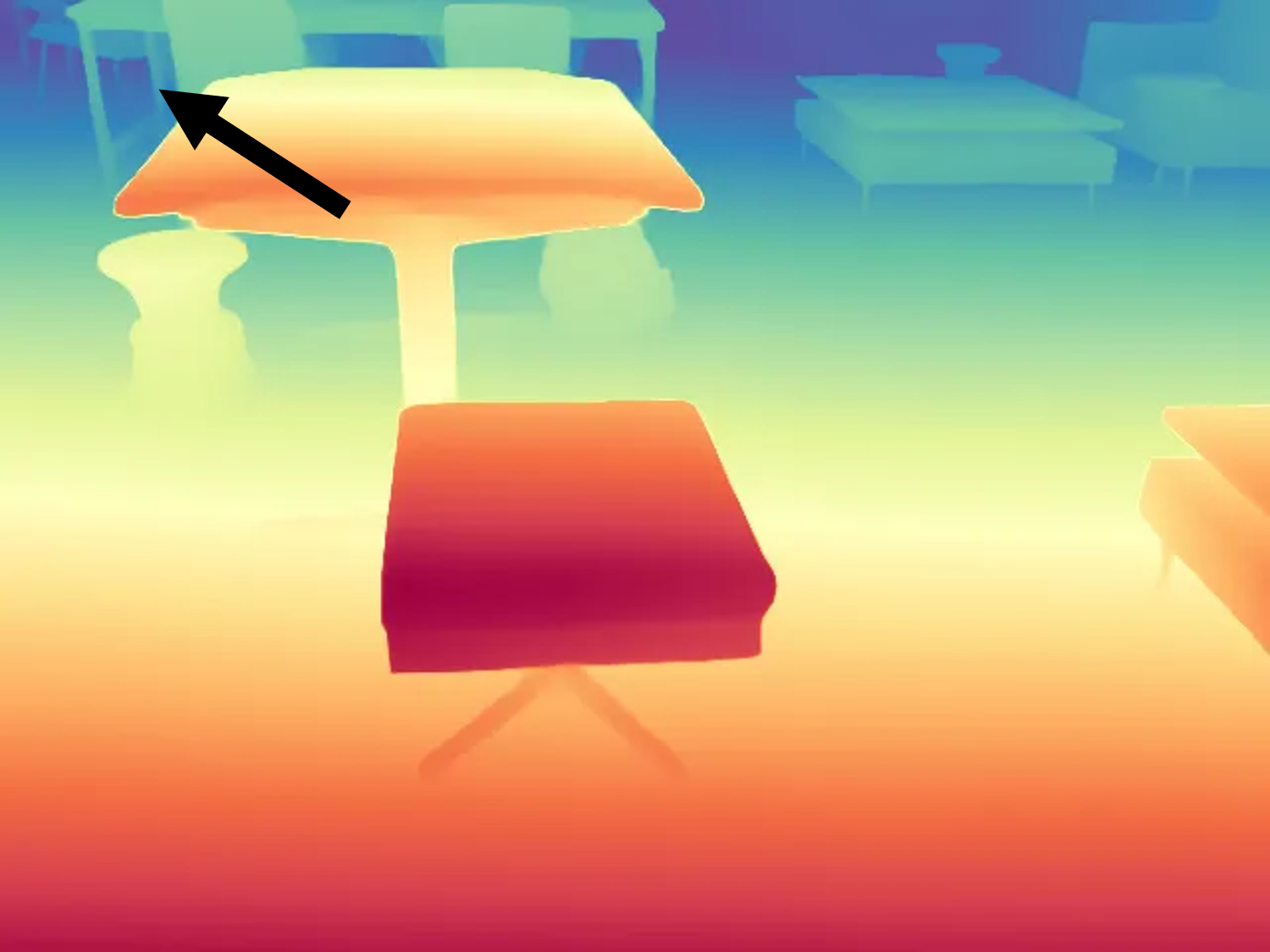}
    \end{subfigure}%
    \hspace{\columnsep}%
    \begin{subfigure}{\imagecolwidth}
    \includegraphics[width=\linewidth]{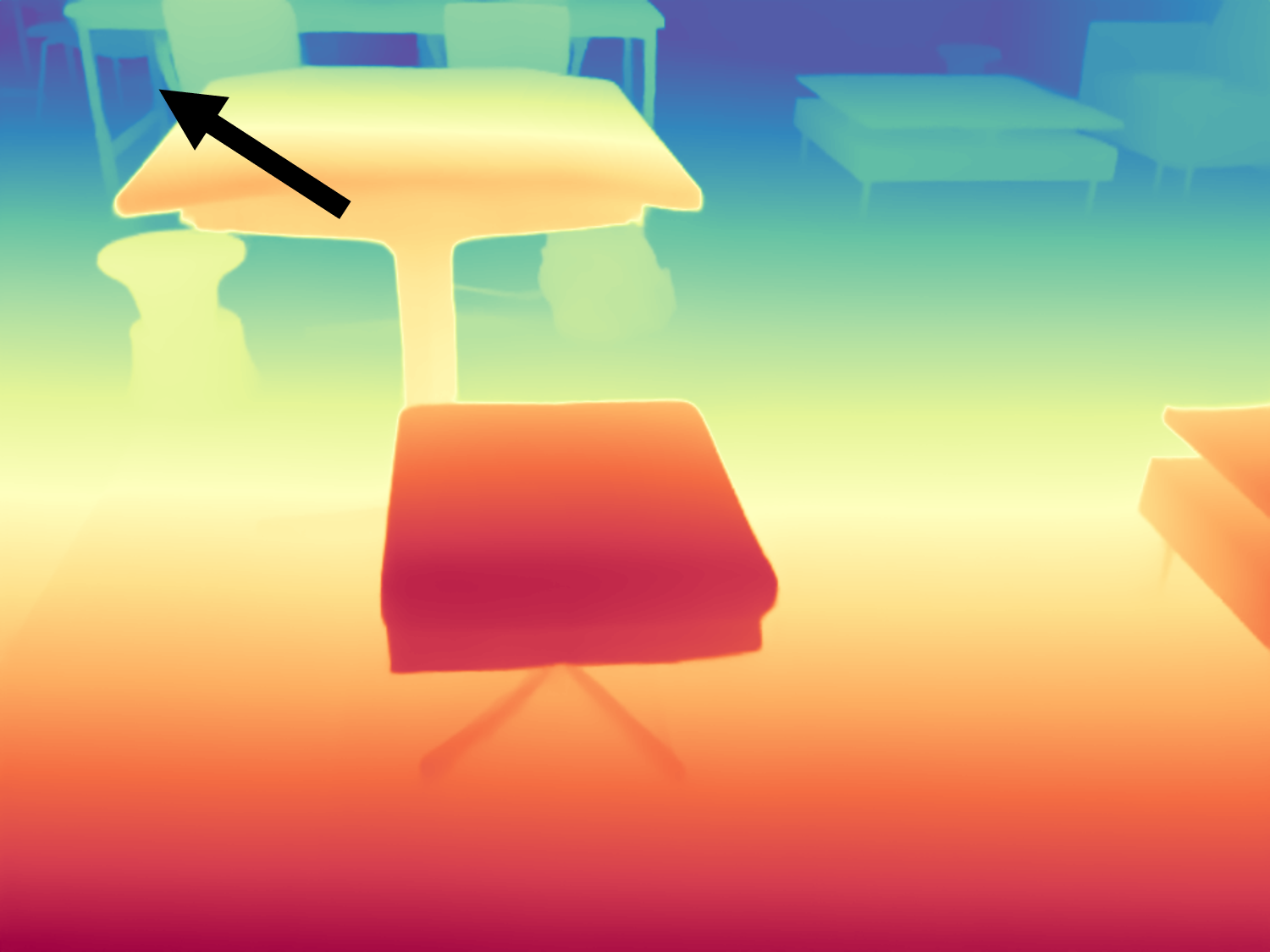} % Placeholder for Depth Pro image
    \end{subfigure}%
    \hspace{\columnsep}%
    \begin{subfigure}{\imagecolwidth}
    \includegraphics[width=\linewidth]{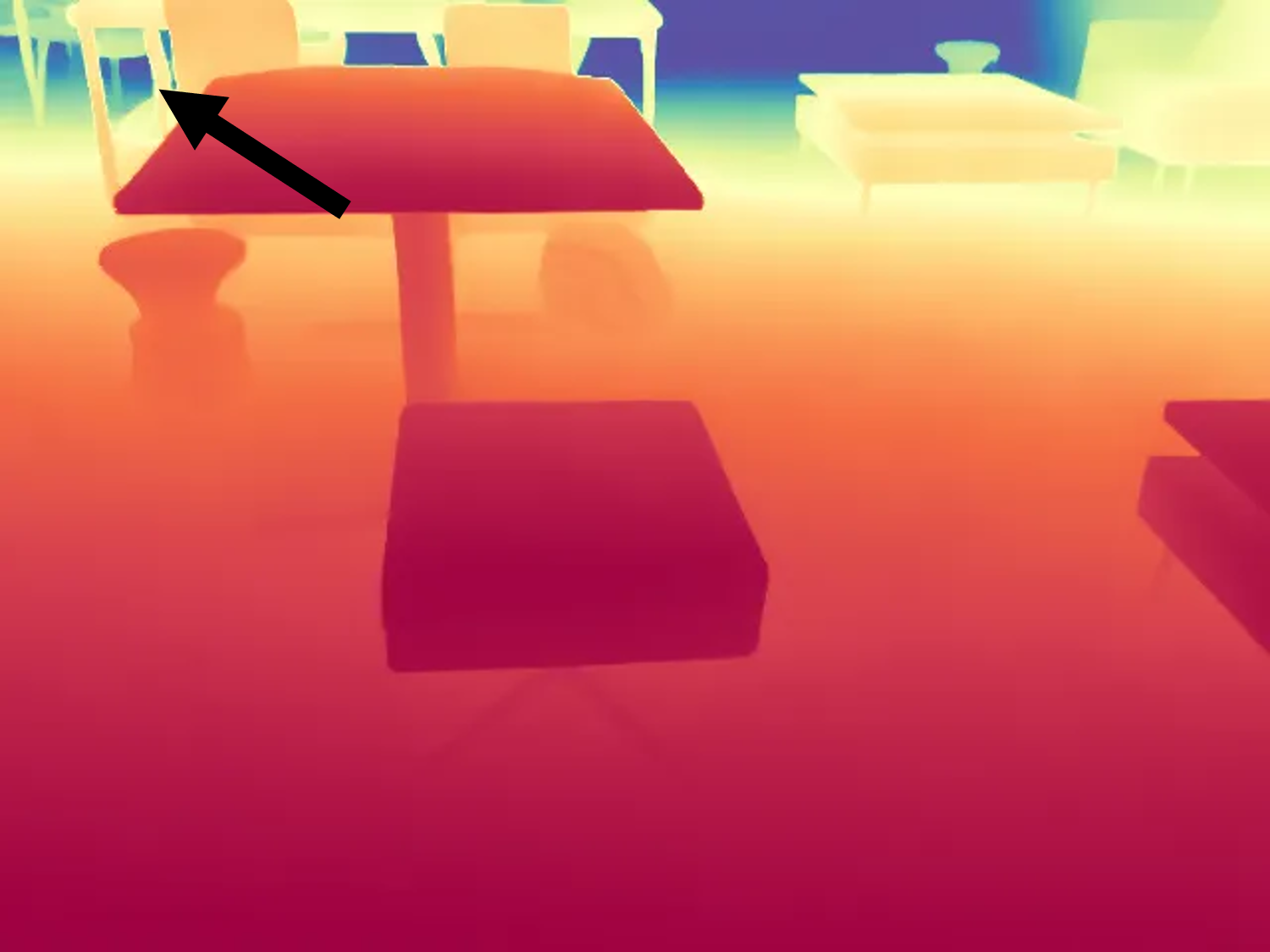}
    \end{subfigure}%
    \hspace{\columnsep}%
    \begin{subfigure}{\imagecolwidth}
    \includegraphics[width=\linewidth]{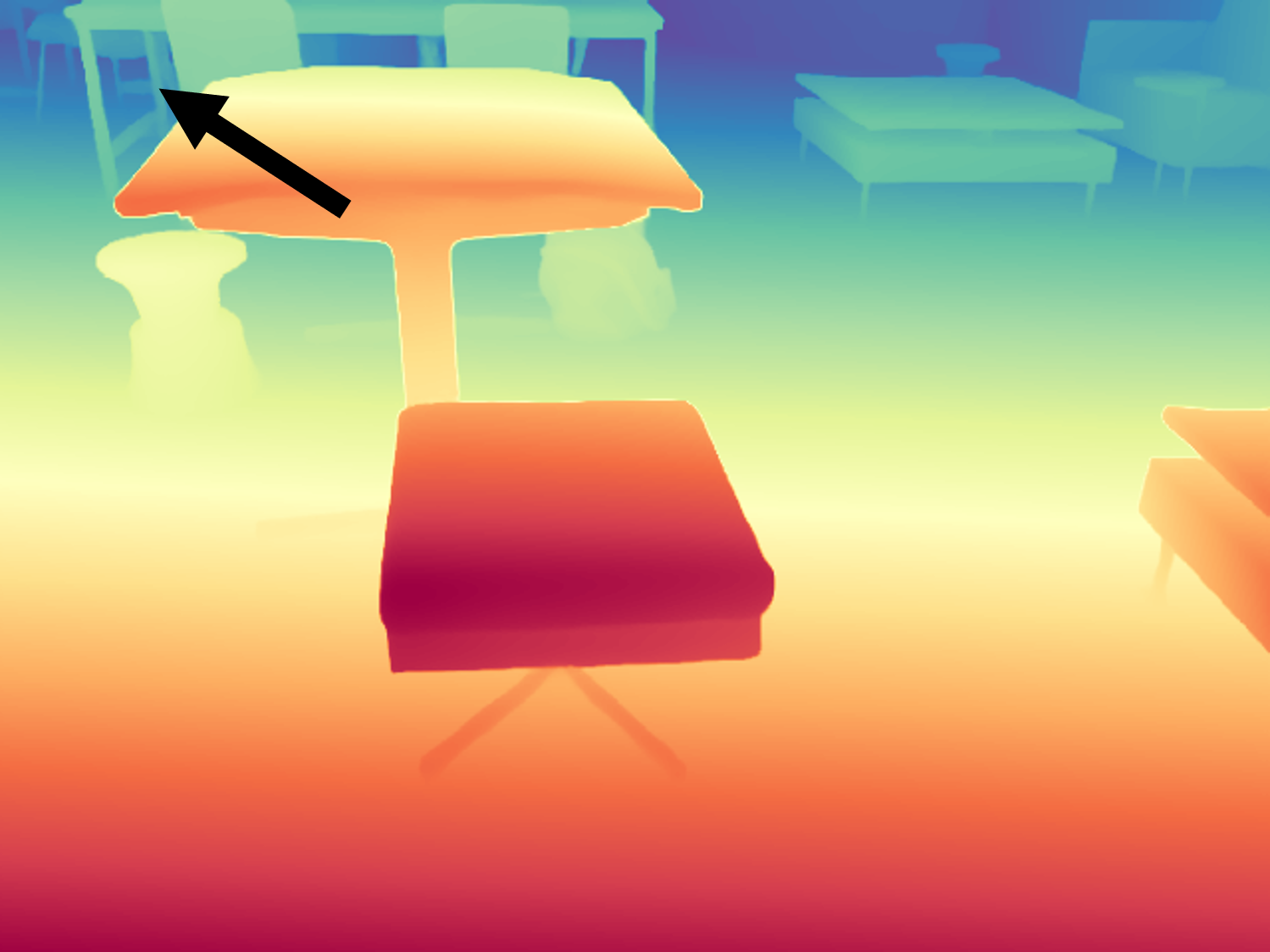}
    \end{subfigure}\\
    
    % Fourth row: ETH3D
    \raisebox{0.4\height}{\makebox[\rowlabelcolwidth][l]{\rotatebox{90}{\small ETH3D}}}%
    \begin{subfigure}{\imagecolwidth}
    \includegraphics[width=\linewidth]{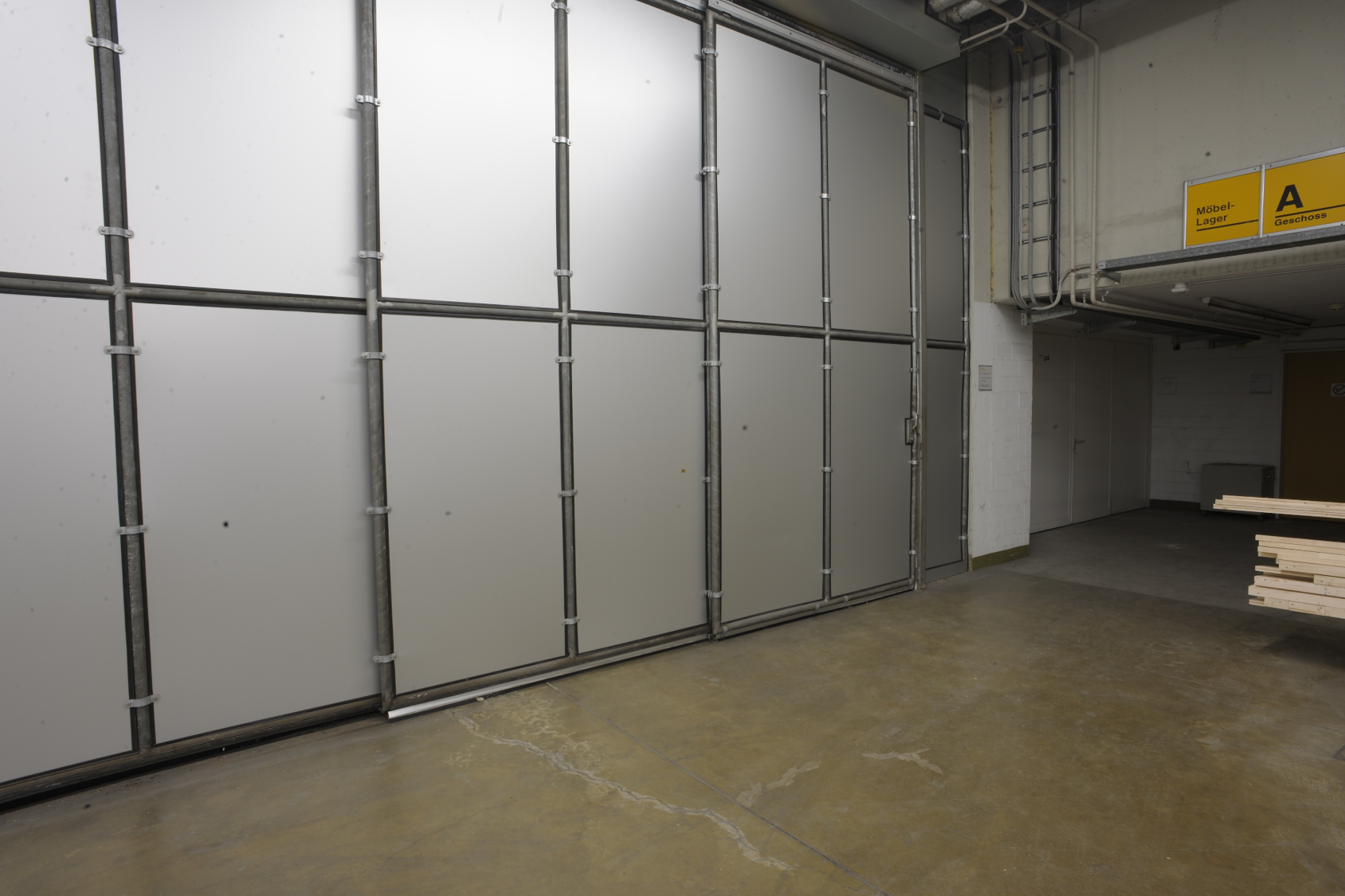}
    \end{subfigure}%
    \hspace{\columnsep}%
    \begin{subfigure}{\imagecolwidth}
    \includegraphics[width=\linewidth]{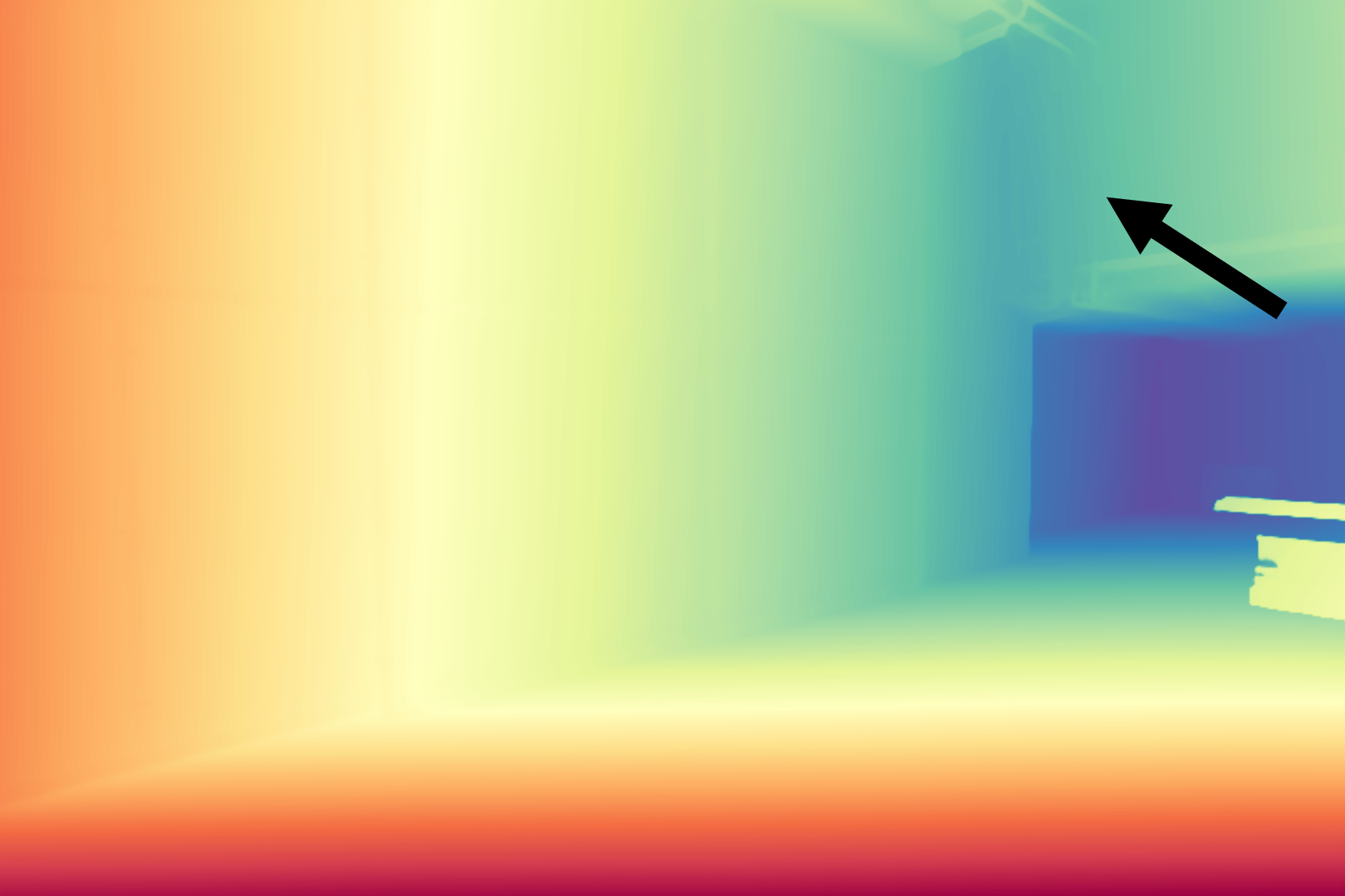}
    \end{subfigure}%
    \hspace{\columnsep}%
    \begin{subfigure}{\imagecolwidth}
    \includegraphics[width=\linewidth]{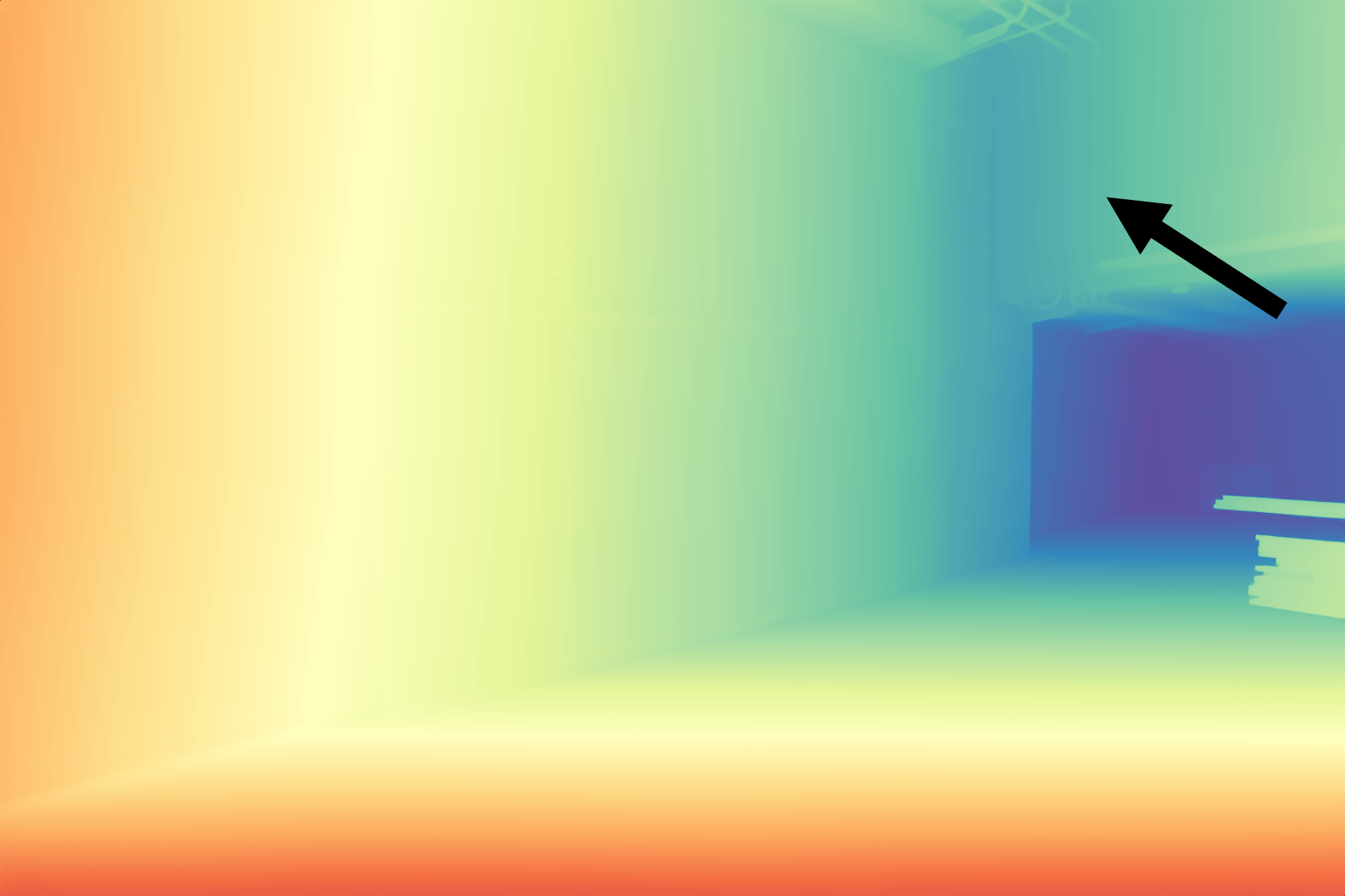} % Placeholder for Depth Pro image
    \end{subfigure}%
    \hspace{\columnsep}%
    \begin{subfigure}{\imagecolwidth}
    \includegraphics[width=\linewidth]{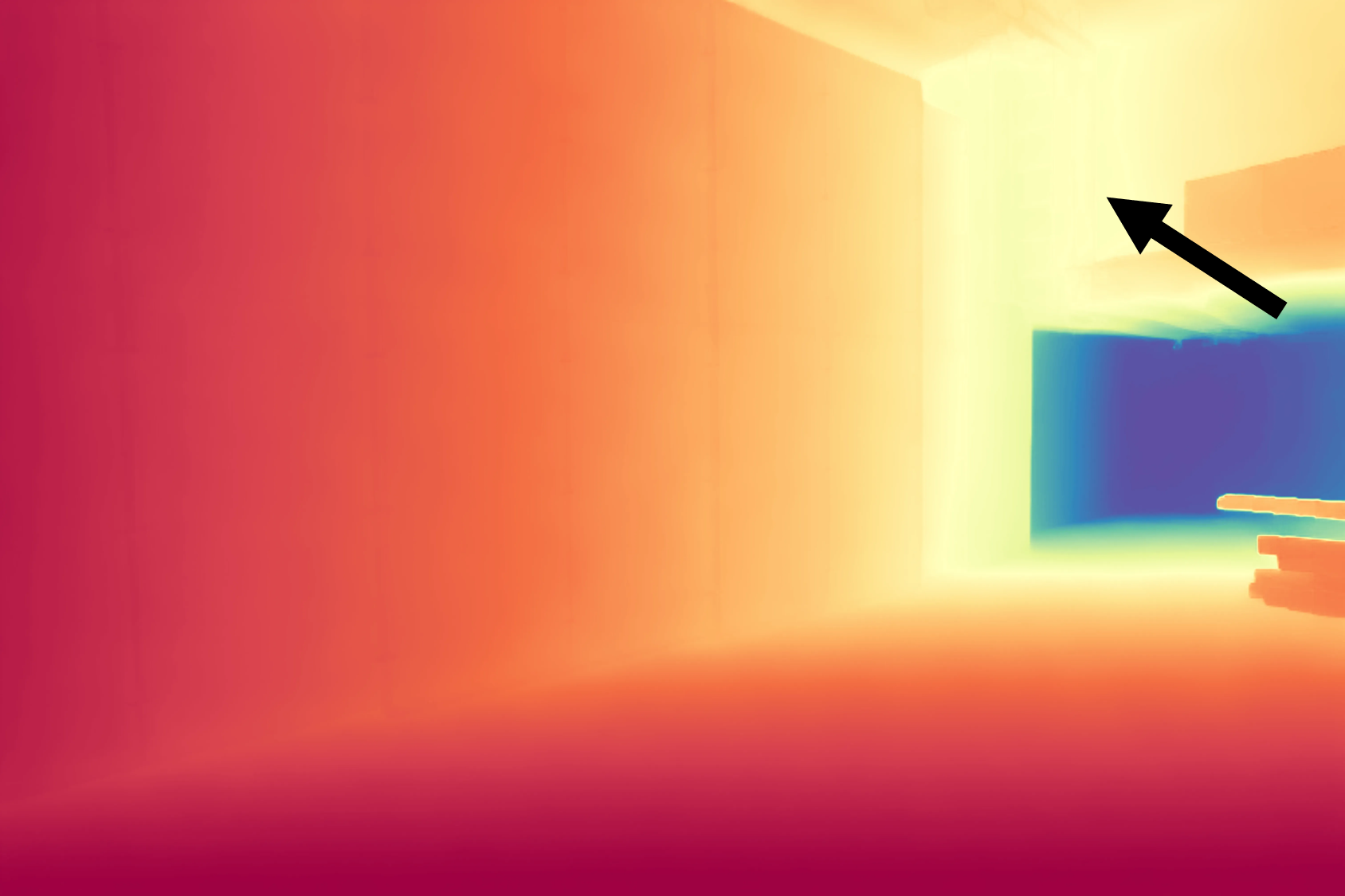}
    \end{subfigure}%
    \hspace{\columnsep}%
    \begin{subfigure}{\imagecolwidth}
    \includegraphics[width=\linewidth]{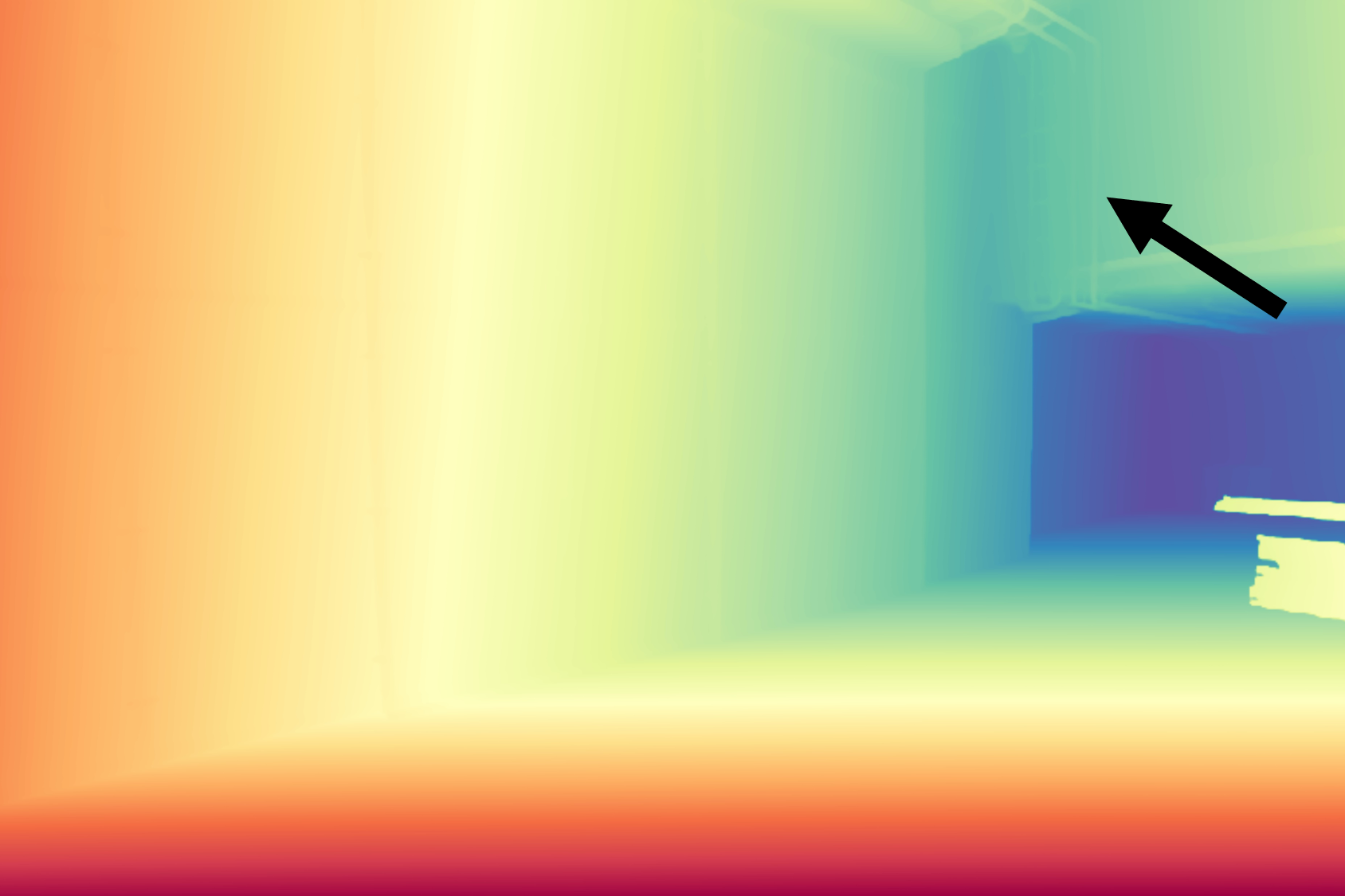}
    \end{subfigure}\\
    
    % Fifth row: Sintel
    \raisebox{0.3\height}{\makebox[\rowlabelcolwidth][l]{\rotatebox{90}{\small Sintel}}}%
    \begin{subfigure}{\imagecolwidth}
    \includegraphics[width=\linewidth]{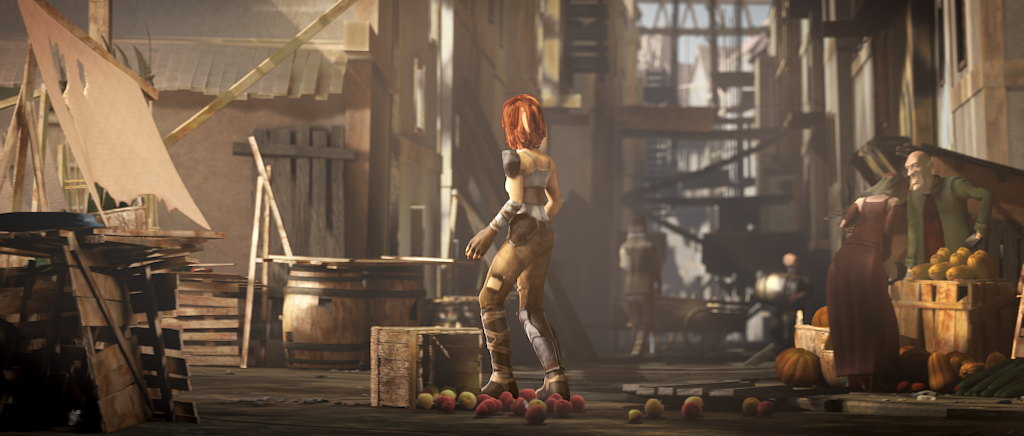}
    \end{subfigure}%
    \hspace{\columnsep}%
    \begin{subfigure}{\imagecolwidth}
    \includegraphics[width=\linewidth]{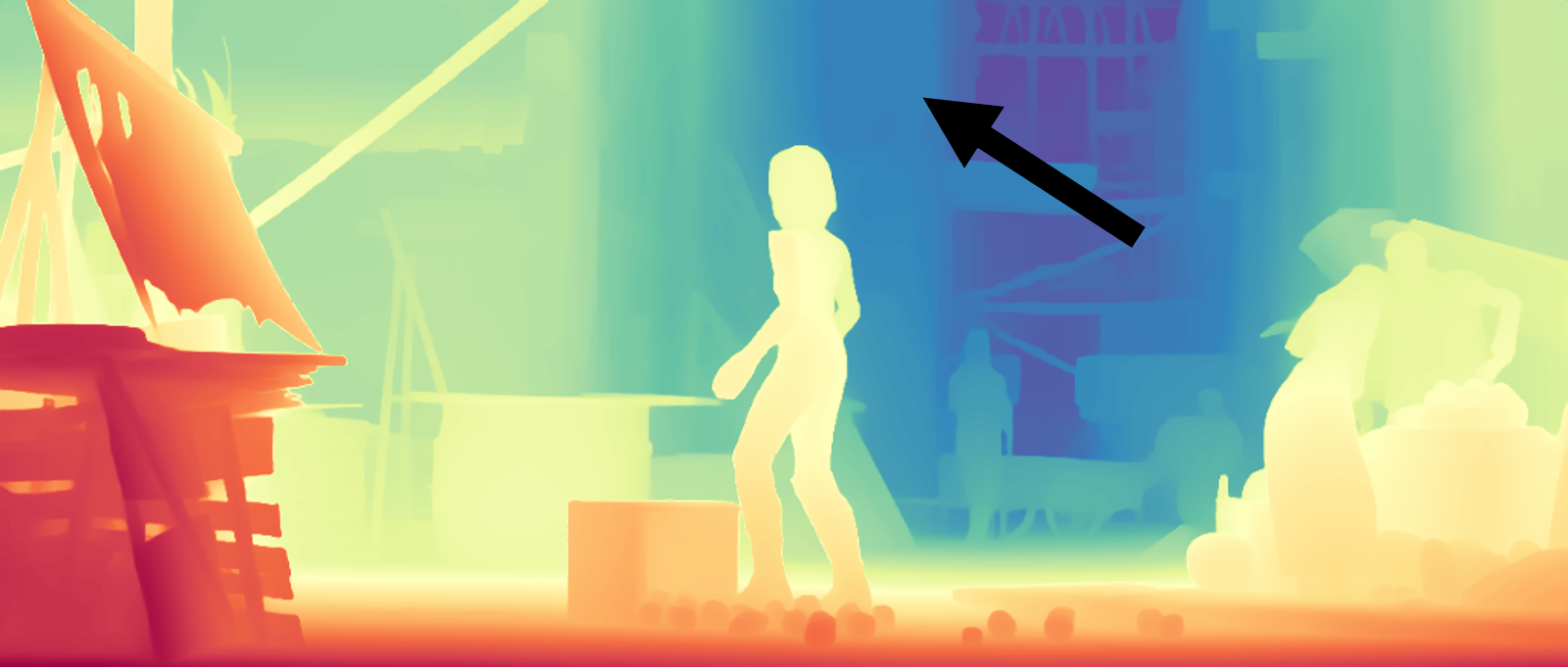}
    \end{subfigure}%
    \hspace{\columnsep}%
    \begin{subfigure}{\imagecolwidth}
    \includegraphics[width=\linewidth]{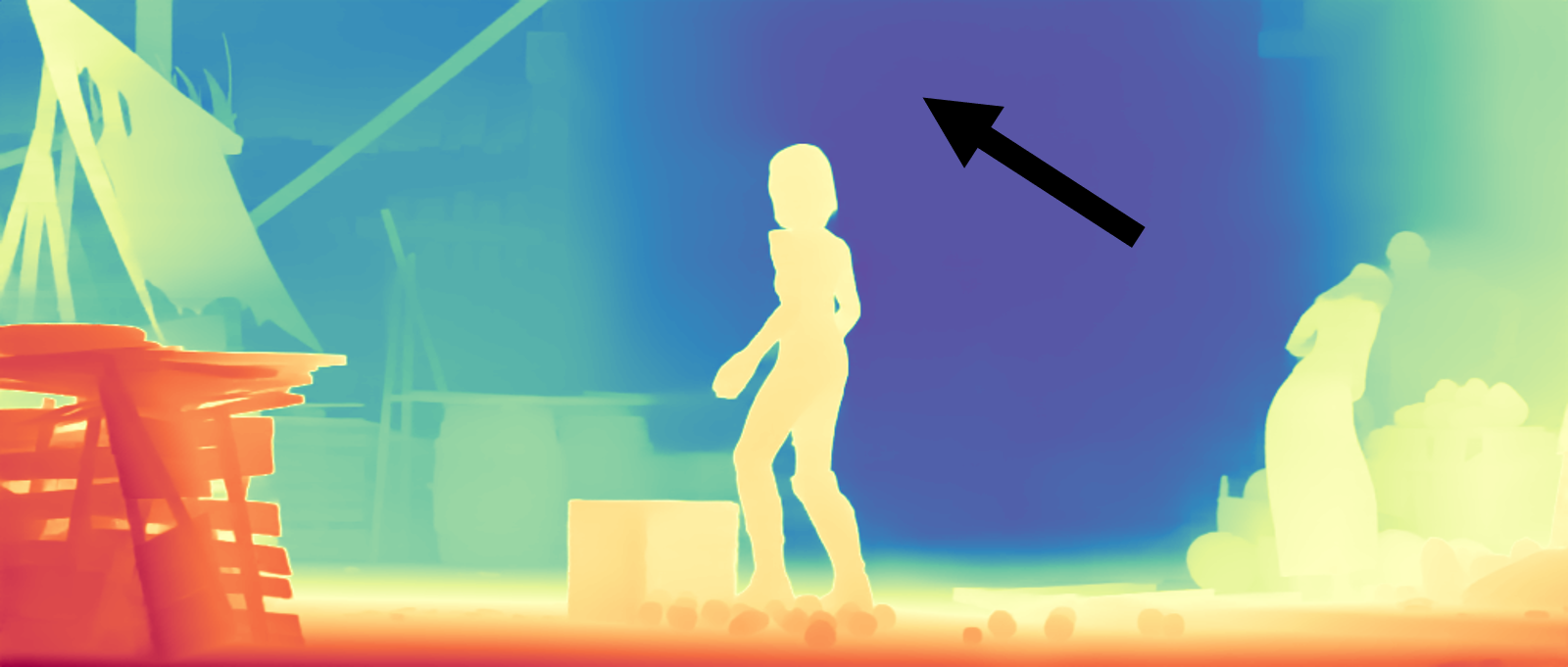} 
    \end{subfigure}%
    \hspace{\columnsep}%
    \begin{subfigure}{\imagecolwidth}
    \includegraphics[width=\linewidth]{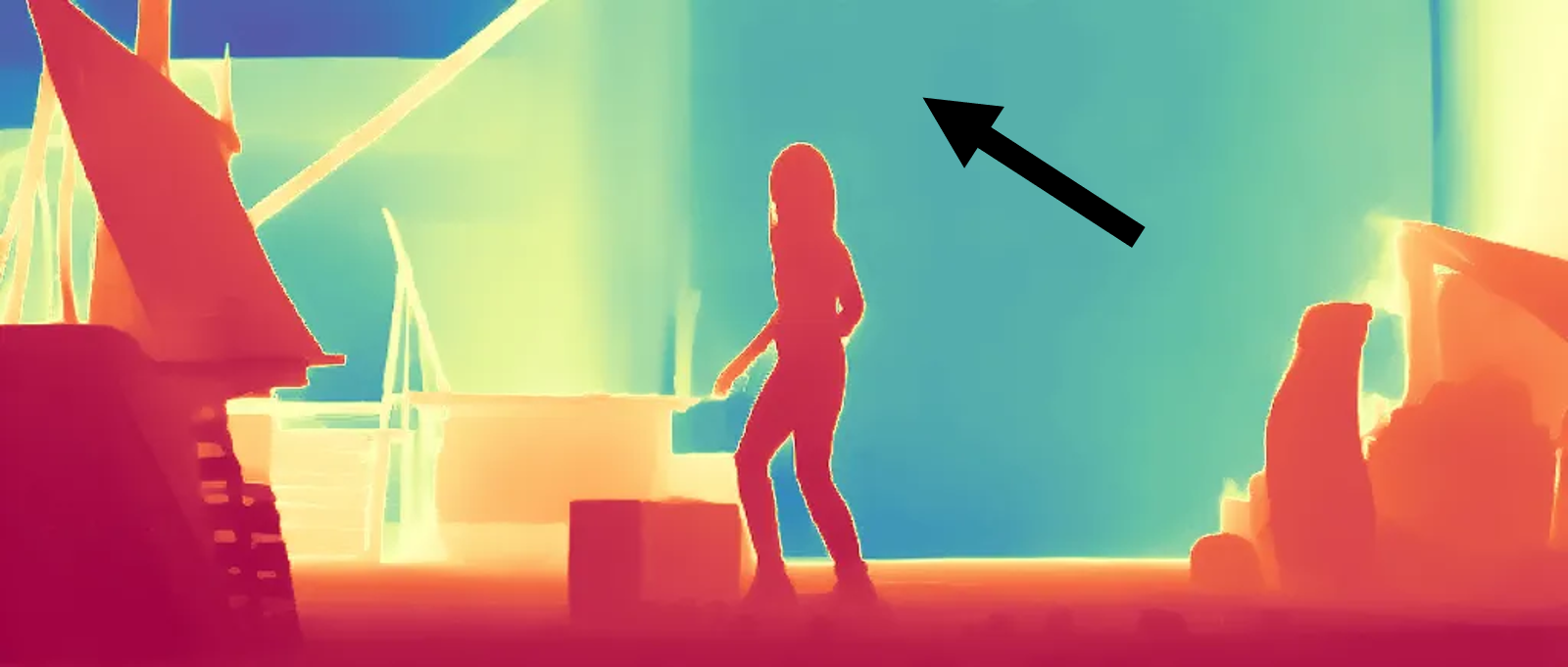}
    \end{subfigure}%
    \hspace{\columnsep}%
    \begin{subfigure}{\imagecolwidth}
    \includegraphics[width=\linewidth]{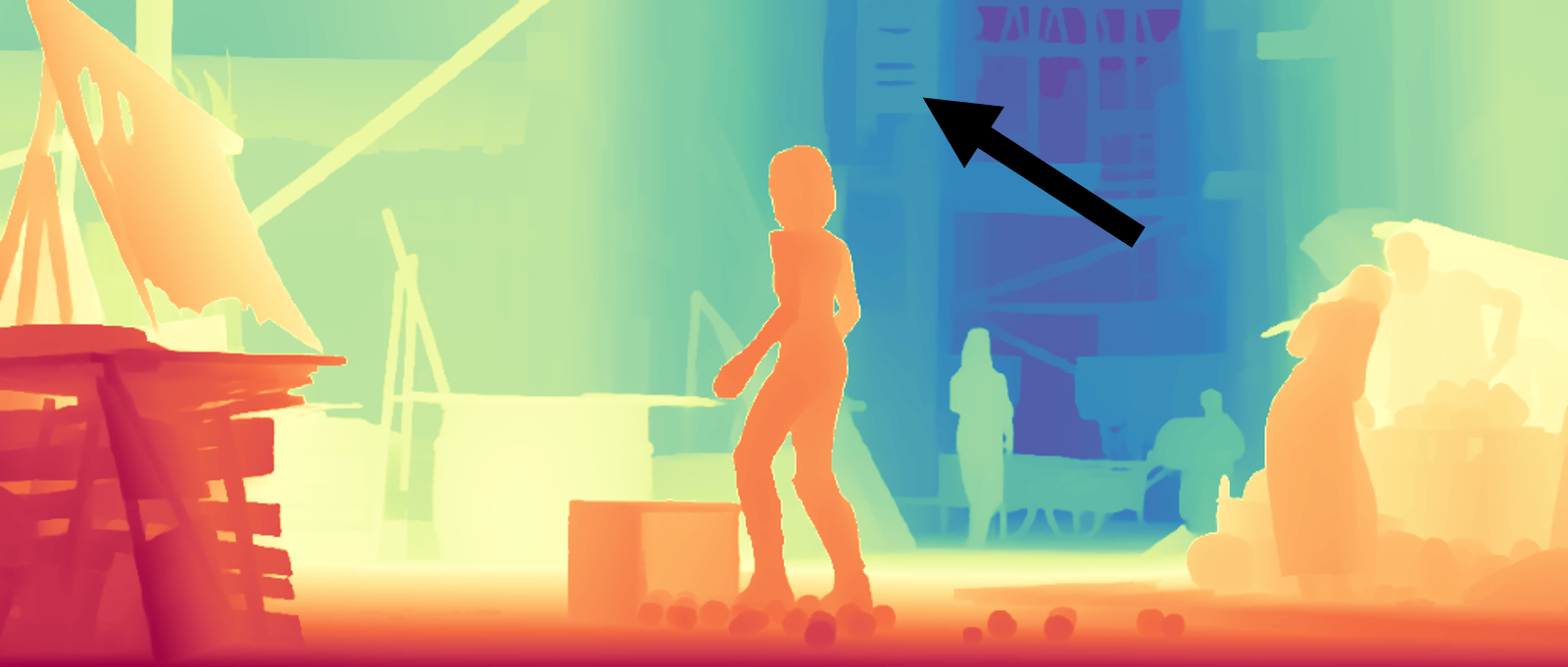}
    \end{subfigure}\\
    
    % Sixth row: In the wild
    \raisebox{0.2\height}{\makebox[\rowlabelcolwidth][l]{\rotatebox{90}{\small \mbox{In the wild}}}}%
    \begin{subfigure}{\imagecolwidth}
    \raisebox{0.5mm}{\includegraphics[width=\linewidth]{Figs/Fig3/rgb6.png}} 
    \end{subfigure}% <-- 修复：添加了%以避免额外空格
    \hspace{\columnsep}%
    \begin{subfigure}{\imagecolwidth}
    \includegraphics[width=\linewidth]{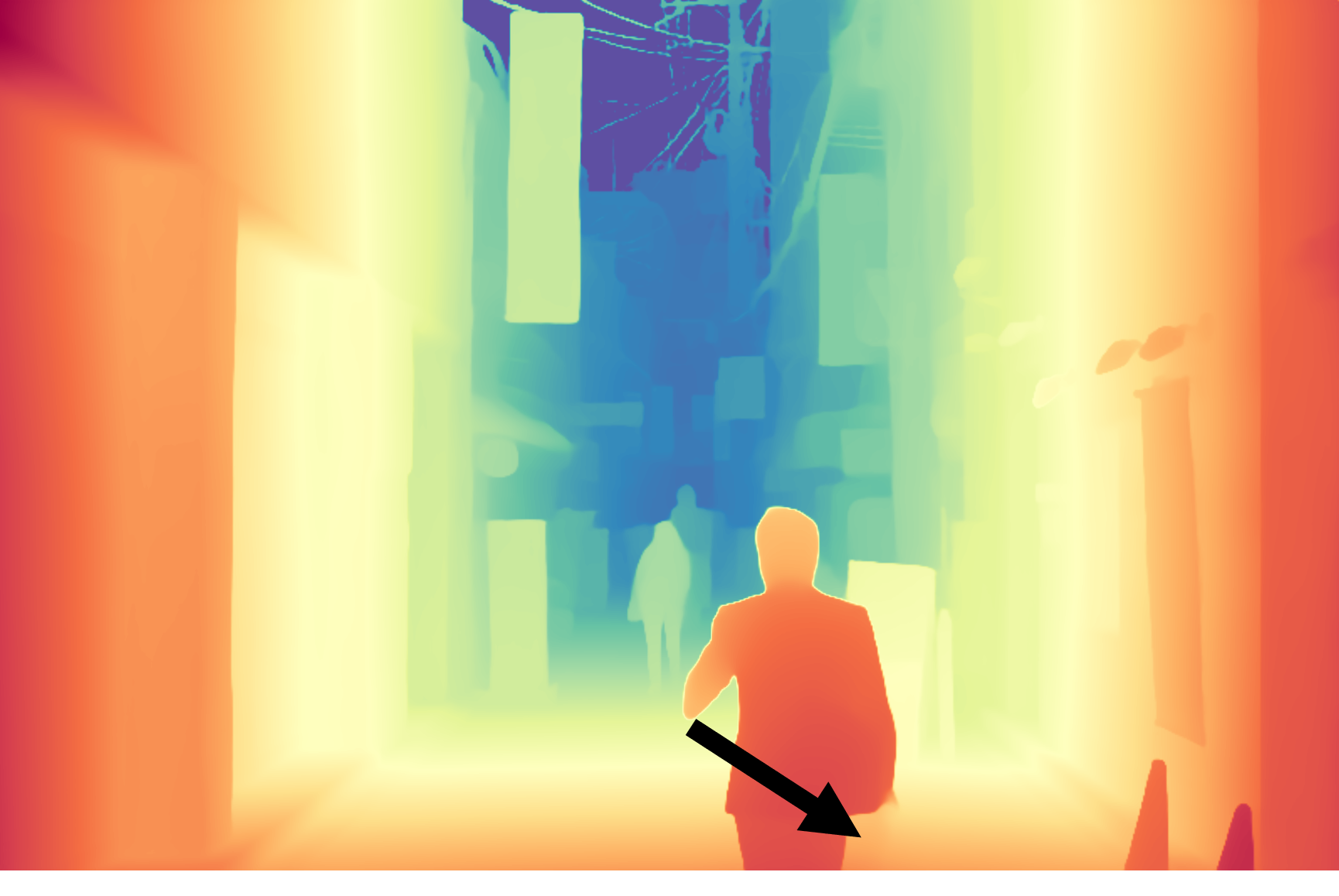}
    \end{subfigure}%
    \hspace{\columnsep}%
    \begin{subfigure}{\imagecolwidth}
    \includegraphics[width=\linewidth]{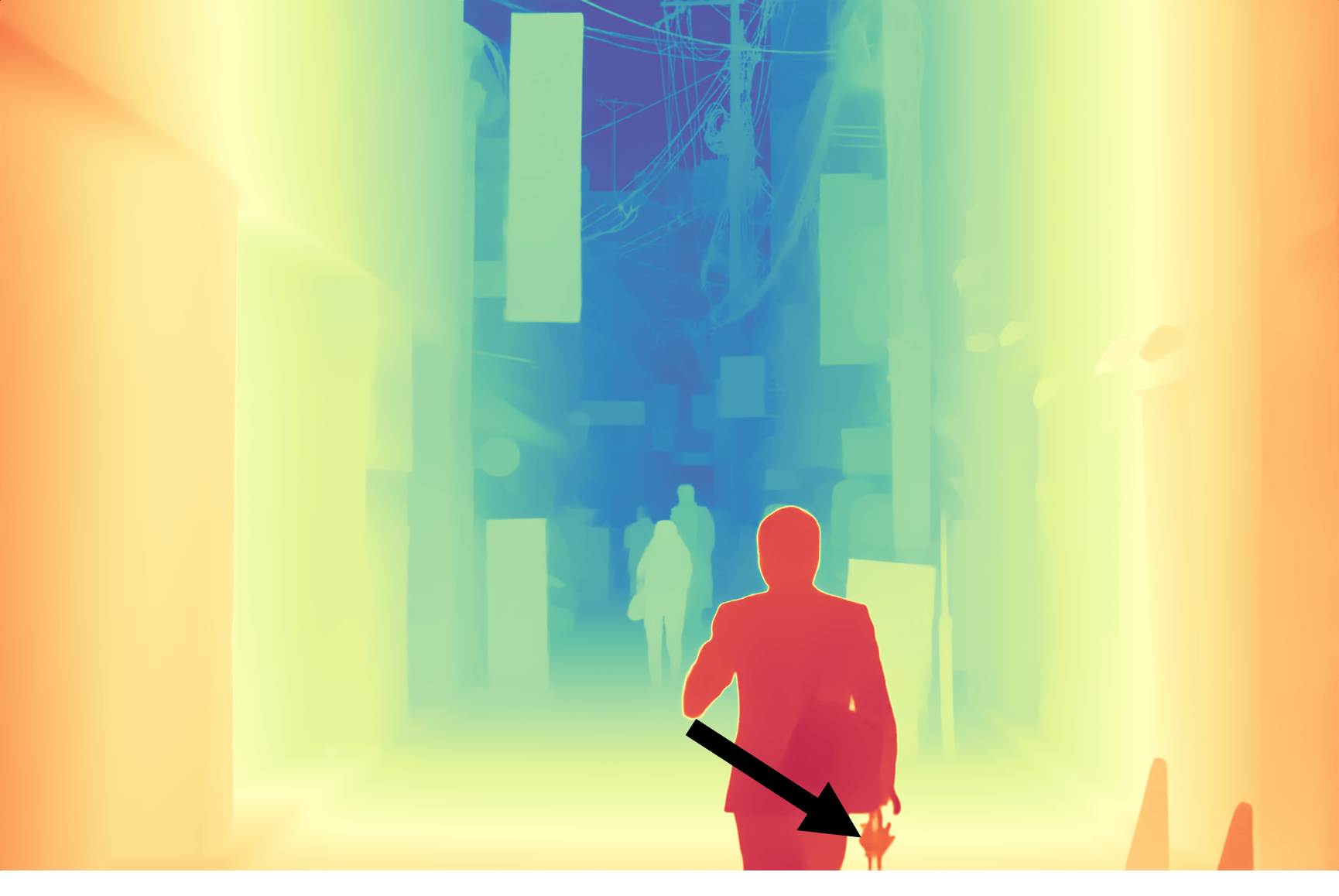}
    \end{subfigure}%
    \hspace{\columnsep}%
    \begin{subfigure}{\imagecolwidth}
    \includegraphics[width=\linewidth]{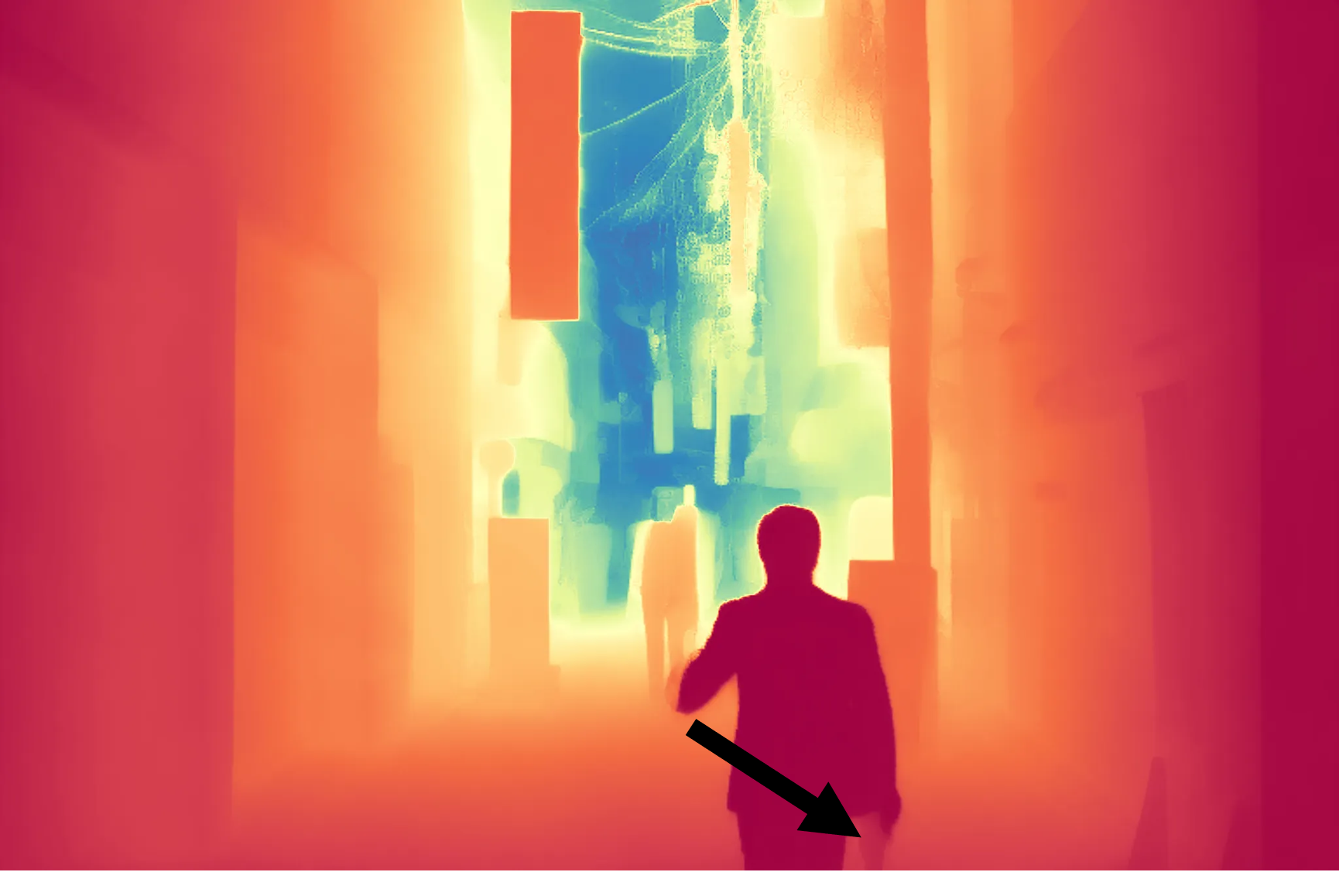}
    \end{subfigure}%
    \hspace{\columnsep}%
    \begin{subfigure}{\imagecolwidth}
    \includegraphics[width=\linewidth]{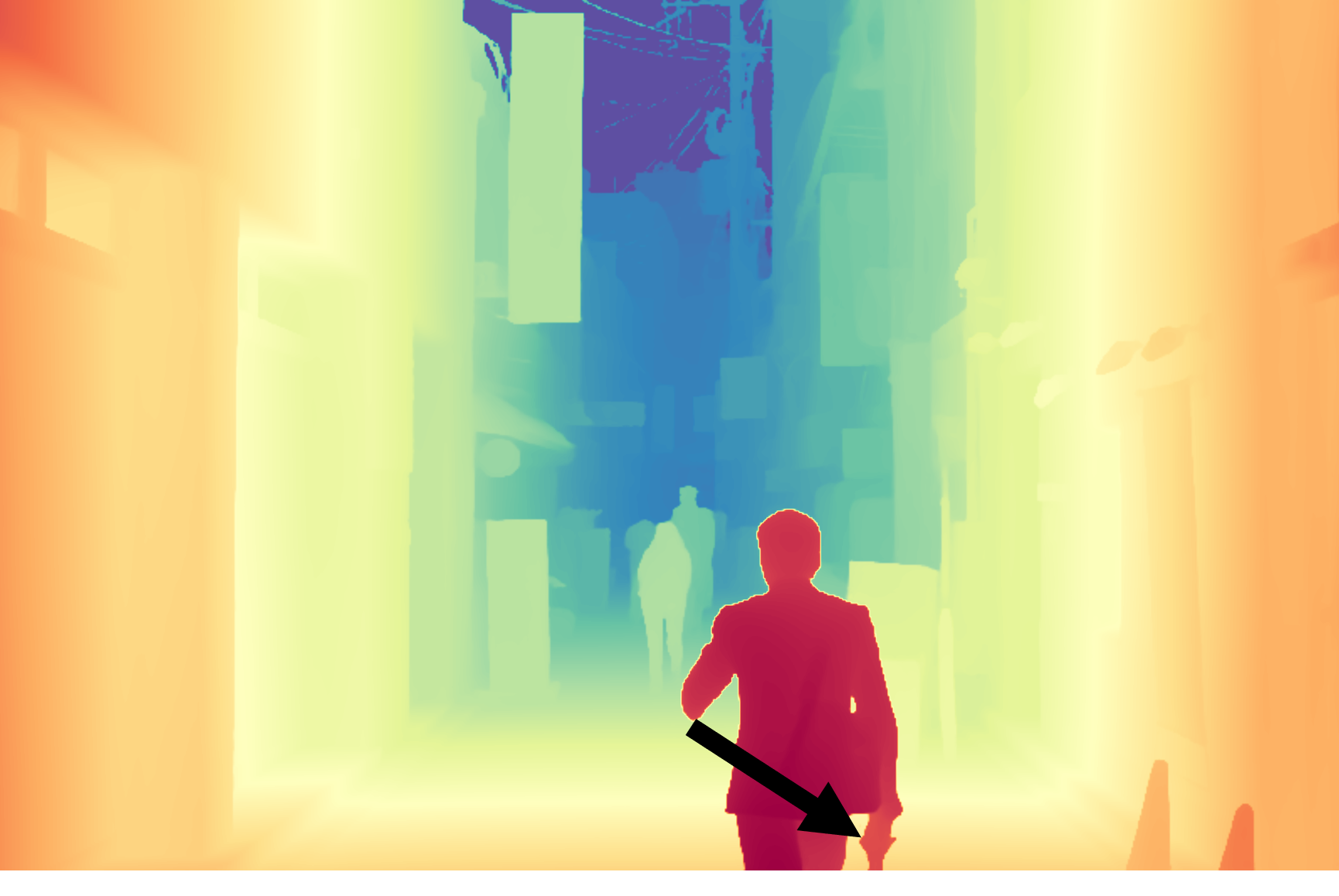}
    \end{subfigure}% 
    
    \caption{\textbf{Qualitative comparison of relative depth estimation across different datasets.} Our model captures fine-grained details and generalizes robustly on in-the-wild data. }
    \label{fig:qualitative}
    \end{figure}
}

\textbf{Quantitative Comparisons:} Table~\ref{tab:relative_depth} shows the results that our method achieves outstanding performance across multiple datasets. Our method achieves outstanding performance across multiple datasets and establishes new state-of-the-art (SOTA) results on several mainstream benchmarks. Specifically, our model demonstrates overwhelming superiority on indoor scene datasets like NYUv2, ScanNet, and ETH3D, with its ability to generate fine-structure predictions aligning perfectly with its objectives. Furthermore, our model slightly surpasses Depth Anything V2 on benchmarks like DA2K, further confirming its excellent zero-shot generalization capabilities. Notably, Depth Pro's significantly lower performance in this evaluation is related to its primary optimization for metric depth rather than relative depth. Lastly, we do not achieve optimal performance on KITTI, primarily because of the dataset's inherent sparsity. Our model is designed to capture fine-grained global and local depth information, which is not fully reflected in the KITTI evaluation.

\textbf{Qualitative Comparisons:} Figure~\ref{fig:qualitative} presents qualitative monocular depth estimation results, highlighting our model's superior capability to capture fine-grained details and robustly handle challenging objects compared to Depth Anything V2, Depth Pro, and generative methods. Notably, our model successfully identifies and accurately estimates the depth of reflective surfaces, such as mirrors, on the NYUv2 dataset. It also preserves complex details, like distant table legs in ScanNet, and captures the fine textures of remote architectural and wooden structures on the Sintel dataset. Furthermore, it effectively processes similarly colored objects; for instance, on the KITTI dataset, it precisely delineates a person's head from the background, a nuance Depth Anything V2 often struggles with. Crucially, our model exhibits excellent generalization to ``in-the-wild'' data, exemplified by its accurate depth estimation for transparent umbrellas, fully demonstrating its robustness in complex real-world scenarios.

\subsection{Fine-tuned to Metric Depth Estimation}
To validate our model's generalization ability in metric depth estimation, we also transfer our pre-trained encoder and design an additional scale head to adapt to both indoor and outdoor domains. The scale head is designed to validate the model's generalization ability in metric depth estimation, adapting to both indoor and outdoor domains. Structurally, it uses the DINOv2 encoder and an MLP head to predict an accurate metric scale. Following the pipeline of Depth Anything V2, our model maintains a consistently strong performance level on both NYUv2 and KITTI datasets shown in Table~\ref{tab:nyu_results} and Table~\ref{tab:kitti_results}.

\subsection{Ablation Studies}
In this section, we systematically evaluate the contribution of each factor in BRIDGE to the overall performance, aiming to quantify the impact of our proposed innovative data generation scheme. Our ablation experiments are conducted under consistent training configurations to ensure fair comparisons.
\begin{table}[ht]
    \caption{\textbf{Impact of Generated Data.} We demonstrate the effectiveness of our RL-D2I engine by comparing a model trained on standard synthetic data with one trained on 1M generated data with ViT-L. The results show that our generated data boosts performance across all metrics.}
    \centering
    \resizebox{\linewidth}{!}{
    \setlength{\tabcolsep}{0.5mm} 
    \renewcommand{\arraystretch}{1.5}

    \begin{tabular}{l c@{\hskip 2pt}c c@{\hskip 2pt}c c@{\hskip 2pt}c c@{\hskip 2pt}c c@{\hskip 2pt}c} % Removed one 'c' for DA2K column
         \toprule
         \multirow{2}{*}[-14pt]{Method} & \multicolumn{2}{c}{KITTI} & \multicolumn{2}{c}{NYUv2} & \multicolumn{2}{c}{ScanNet} & \multicolumn{2}{c}{ETH3D} & \multicolumn{2}{c}{Sintel} \\ % Removed DA2K from header
         \cmidrule(lr){2-3} \cmidrule(lr){4-5} \cmidrule(lr){6-7} \cmidrule(lr){8-9} \cmidrule(lr){10-11} % Adjusted cmidrule indices
         & $\delta_1\uparrow$ & AbsRel$\downarrow$ & $\delta_1\uparrow$ & AbsRel$\downarrow$ & $\delta_1\uparrow$ & AbsRel$\downarrow$ & $\delta_1\uparrow$ & AbsRel$\downarrow$ & $\delta_1\uparrow$ & AbsRel$\downarrow$ \\
         \midrule 
         Baseline & 0.921 & 0.097 & 0.960 & 0.060 & 0.964 & 0.057 & 0.969 & 0.060 & 0.599 & 0.709 \\ 

         \textbf{w/~Generated Data} & \textbf{0.929} & \textbf{0.092} & \textbf{0.964} & \textbf{0.056} & \textbf{0.968} & \textbf{0.048} & \textbf{0.970} & \textbf{0.059} & \textbf{0.612} & \textbf{0.692} \\
         \bottomrule 
    \end{tabular}
    } 

    \label{tab:ablation_rl_d2i_impact}
\end{table}
% ---------------------
\begin{table}[ht]
    \caption{\textbf{Ablation Study of the Hybrid Training Strategy.} We compares the performance of BRIDGE's ViT-L model, trained solely on pseudo-labels with hybrid supervision integrating similarity-guided high-precision ground truth depth on benchmarks. The results highlight the importance of incorporating original ground truth depth for improving depth estimation accuracy and detail capture.}
    \centering
    \resizebox{\linewidth}{!}{
    \setlength{\tabcolsep}{0.5mm} 
    \renewcommand{\arraystretch}{1.5}
    \begin{tabular}{l c@{\hskip 2pt}c c@{\hskip 2pt}c c@{\hskip 2pt}c c@{\hskip 2pt}c c@{\hskip 2pt}c}
         \toprule
         \multirow{2}{*}[-14pt]{Method} & \multicolumn{2}{c}{KITTI} & \multicolumn{2}{c}{NYUv2} & \multicolumn{2}{c}{ScanNet} & \multicolumn{2}{c}{ETH3D} & \multicolumn{2}{c}{Sintel} \\
         \cmidrule(lr){2-3} \cmidrule(lr){4-5} \cmidrule(lr){6-7} \cmidrule(lr){8-9} \cmidrule(lr){10-11}
         & $\delta_1\uparrow$ & AbsRel$\downarrow$ & $\delta_1\uparrow$ & AbsRel$\downarrow$ & $\delta_1\uparrow$ & AbsRel$\downarrow$ & $\delta_1\uparrow$ & AbsRel$\downarrow$ & $\delta_1\uparrow$ & AbsRel$\downarrow$ \\
         \midrule 
         Pseudo-labels Only & 0.924 & 0.099 & 0.958 & 0.069 & 0.957 & 0.069 & 0.949 & 0.078 & 0.595 & 0.751 \\
         \textbf{Full Hybrid Strategy} & \textbf{0.926} & \textbf{0.098} & \textbf{0.960} & \textbf{0.067} & \textbf{0.960} & \textbf{0.061} & \textbf{0.953} & \textbf{0.073} & \textbf{0.600} & \textbf{0.725} \\

         \bottomrule 
    \end{tabular}
    } 

    \label{tab:rl_d2i_efficiency}
\end{table}

\afterpage{
\begin{table*}[ht]
    \centering
    % 可以稍微增加列间距，因为现在表格内容会被强制缩放
    \setlength{\tabcolsep}{1pt} 
    \captionsetup[sub]{font=small, labelfont=bf}

    % 第一个子表格 (a) for NYU-D dataset
    \begin{minipage}{0.48\textwidth} 
        \caption{The metric results on NYUv2.}
        \centering
        \resizebox{\linewidth}{!}{ % 强制表格内容缩放到minipage的宽度
        \begin{tabular}{l cc cc}
            
            \toprule
            \multirow{2}{*}[-10pt]{Method} & \multicolumn{1}{c}{Higher $\uparrow$} & \multicolumn{2}{c}{Lower $\downarrow$} \\
            \cmidrule(lr){2-2} \cmidrule(lr){3-4}
            & $\delta_1$ & AbsRel & RMSE \\
            \midrule
            AdaBins~\citep{bhat2021adabins} & 0.903 & 0.103 & 0.364 \\
            DPT~\citep{ranftl2021vision} & 0.904 & 0.110 & 0.357 \\
            P3Depth~\citep{patil2022p3depth} & 0.898 & 0.104 & 0.356 \\
            SwinV2~\citep{liu2022swin} & 0.949 & 0.083 & 0.287 \\
            IEBins~\citep{shao2023iebins} & 0.936 & 0.087 & 0.314 \\
            ZoeDepth~\citep{zoedepth} & 0.951 & 0.077 & 0.282 \\
            Metric3Dv2~\citep{hu2024metric3d} & 0.987 & 0.046 & 0.181 \\
            \midrule
            Ours & \textbf{0.986} & \textbf{0.052} & \textbf{0.197} \\
            \bottomrule
        \end{tabular}
        
        } 
        \label{tab:nyu_results}
    \end{minipage}
    \hfill % 在两个minipage之间加入灵活的水平空间
    \begin{minipage}{0.48\textwidth} 
        \caption{The metric results on KITTI.}
        \centering
        \resizebox{\linewidth}{!}{ % 强制表格内容缩放到minipage的宽度
        \begin{tabular}{l cc cc}
            
            \toprule
            \multirow{2}{*}[-10pt]{Method} & \multicolumn{1}{c}{Higher $\uparrow$} & \multicolumn{2}{c}{Lower $\downarrow$} \\
            \cmidrule(lr){2-2} \cmidrule(lr){3-4}
            & $\delta_1$ & AbsRel & RMSE \\
            \midrule
            AdaBins~\citep{bhat2021adabins} & 0.964 & 0.058 & 2.360 \\
            P3Depth~\citep{patil2022p3depth} & 0.953 & 0.071 & 2.842 \\
            SwinV2~\citep{liu2022swin} & 0.977 & 0.050 & 1.966 \\
            GEDepth~\citep{yang2023gedepth} & 0.976 & 0.048 & 2.044 \\
            IEBins~\citep{shao2023iebins} & 0.978 & 0.050 & 2.011 \\
            ZoeDepth~\citep{zoedepth} & 0.971 & 0.054 & 2.281 \\
            Metric3Dv2~\citep{hu2024metric3d}  & 0.985 & 0.044 & 1.993 \\
            \midrule
            Ours & \textbf{0.983} & \textbf{0.045} & \textbf{1.862} \\
            \bottomrule
        \end{tabular}
        
        } 
        \label{tab:kitti_results}
    \end{minipage}
    \label{tab:sota_comparison}
\end{table*}
}

\textbf{Impact of \name~Generated Data:} We first investigate the effectiveness of our RL-D2I generation engine, which serves as the core mechanism for alleviating data scarcity and enhancing training data diversity. To demonstrate its significant impact, we compare a baseline model using DINOv2 Base trained on 1M synthetic data from various large-scale synthetic depth datasets with the BRIDGE model generated 1M data (comprising 0.95M pseudo-labels and 0.05M masked original ground truth depth). As shown in Table~\ref{tab:ablation_rl_d2i_impact}, the integration of RL-D2I-generated data significantly boosts the performance across all evaluated metrics. This improvement underscores the effectiveness of our RL-D2I engine in generating visually realistic and geometrically accurate images, which effectively expands the scale and diversity of the training data.

\textbf{Efficiency of Hybrid Training Strategy:} As shown in Table~\ref{tab:rl_d2i_efficiency}, our hybrid depth supervision strategy is crucial for combining the broad coverage of teacher model pseudo-labels with the precision of similarity-guided high-precision ground truth depth. This approach aims to provide more robust and reliable supervision signals for learning geometric structural knowledge. We conduct an ablation study to quantify the benefits of this dual supervision. We also test on 1M data, the BRIDGE model trained with our full hybrid strategy yields superior results compared to a variant that relies solely on teacher-generated pseudo-labels for the generated RGB images.

%% file: Chapter/5.conclusion.tex
\section{Conclusion}
In this work, we introduce \name, an innovative Monocular Depth Estimation framework that addresses data scarcity and quality issues. This is achieved by leveraging an RL-optimized Depth-to-Image (RL-D2I) engine to generate over 20M RGB-D data, which, along with a hybrid strategy of teacher pseudo-labels and high-precision ground truth, achieves state-of-the-art performance, consistently outperforming methods like Depth Anything V2 on benchmarks with significantly less training data (20M vs. 62M). Our findings highlight \name's superior capability in capturing fine-grained details, ensuring geometric consistency, and demonstrating robust zero-shot generalization to complex in-the-wild scenes, paving the way for more efficient and generalizable MDE solutions.

%% file: Chapter/appendix.tex
\section{Experiments}
\subsection{Model Selection}
We compare the learning capabilities of different DINOv2~\citep{oquab2023dinov2} versions on our hybrid data. For DINOv2-Giant and Large, we select the non-register version as their representative. As shown in Table~\ref{tab:as}, we find that the Giant and Large versions of DINOv2 exhibit excellent learning ability on synthetic data and generalize remarkably well to real-world test datasets. In contrast, the Small and Base versions do not achieve satisfactory results. This might be attributed to their limited model capacity, which is insufficient to capture the rich fine-grained information and diversity present in generation data, thus leading to difficulties in generalizing to more complex real-world scenarios. Therefore, we ultimately choose to adopt the DINOv2-Giant for our model, to fully leverage its powerful generalization capabilities and accuracy.

\begin{table}[ht]
    \centering
    \resizebox{0.8\linewidth}{!}{
    \setlength{\tabcolsep}{0.5mm} 
    \renewcommand{\arraystretch}{1.5}

    \begin{tabular}
    {l c@{\hskip 2pt}c c@{\hskip 2pt}c c@{\hskip 2pt}c c@{\hskip 2pt}c c@{\hskip 2pt}c} % Removed one 'c' for DA2K column
         \toprule
         \multirow{2}{*}[-14pt]{Method} & \multicolumn{2}{c}{KITTI} & \multicolumn{2}{c}{NYUv2} & \multicolumn{2}{c}{ScanNet} & \multicolumn{2}{c}{ETH3D} & \multicolumn{2}{c}{Sintel} \\ % Removed DA2K from header
         \cmidrule(lr){2-3} \cmidrule(lr){4-5} \cmidrule(lr){6-7} \cmidrule(lr){8-9} \cmidrule(lr){10-11} % Adjusted cmidrule indices
         & $\delta_1\uparrow$ & AbsRel$\downarrow$ & $\delta_1\uparrow$ & AbsRel$\downarrow$ & $\delta_1\uparrow$ & AbsRel$\downarrow$ & $\delta_1\uparrow$ & AbsRel$\downarrow$ & $\delta_1\uparrow$ & AbsRel$\downarrow$ \\
         \midrule 
         DINOv2-S & 0.928 & 0.082 & 0.964 & 0.062 & 0.921 & 0.085 & 0.978 & 0.045 & 0.637 & 0.684 \\ 
        DINOv2-B & 0.932 & 0.081 & 0.968 & 0.054 & 0.941 & 0.065 & 0.980 & 0.039 & 0.647 & 0.644 \\
        DINOv2-L & 0.937 & 0.081 & 0.978 & 0.045 & 0.972 & 0.049 & 0.989 & 0.032 & 0.688 & 0.588 \\ 
        DINOv2-G & \textbf{0.938} & \textbf{0.081} & \textbf{0.982} & \textbf{0.041} & \textbf{0.981} & \textbf{0.033} & \textbf{0.991} & \textbf{0.029} & \textbf{0.719} & \textbf{0.513} \\ 
         \bottomrule 
    \end{tabular}
} 
    \caption{Comparison among various size pre-trained DINOv2 encoders trained on our dataset.}

    \label{tab:as}
\end{table}

\begin{table}[ht]
    \caption{\textbf{Our training sources. * represents the quantity sampled from the dataset.}}
    \centering
    \resizebox{0.5\linewidth}{!}{ 
    \setlength{\tabcolsep}{0.5mm} 
    \renewcommand{\arraystretch}{1.5}
    \begin{tabular}{l c c r}
         \toprule
         \textbf{Dataset} & \textbf{Indoor} & \textbf{Outdoor} & \textbf{\# Images} \\
         \midrule 
         \multicolumn{4}{l}{\textbf{Precise Synthetic Images (1M)}} \\ 
         \midrule
         BlendedMVS~\citep{yao2020blendedmvs} & \checkmark & \checkmark & 115K \\
        TartanAir~\citep{wang2020tartanair} & \checkmark & \checkmark & 306K \\
         Hypersim~\citep{roberts2021hypersim} & \checkmark & & 60K \\
         IRS~\citep{wang2019irs} & \checkmark & & 103K \\
         VKITTI 2~\citep{cabon2020virtual} & & \checkmark & 20K \\
          Spring~\citep{mehl2023spring}&  & \checkmark & 5K \\

          DA-V~\citep{yang2024depth} & \checkmark & \checkmark & 400K* \\
         \midrule 
         \multicolumn{4}{l}{\textbf{Generated Images (20M)}} \\ 
         \midrule
         BlendedMVS & \checkmark & \checkmark & 2.2M \\
         TartanAir & & \checkmark & 2M \\
         Hypersim & \checkmark & \checkmark & 2.4M \\
         IRS & \checkmark & \checkmark & 4.2M \\
         VKITTI 2 &  & \checkmark & 1M \\
         Spring &  & \checkmark & 20K \\
         DA-V & \checkmark & \checkmark & 8.2M \\
         \bottomrule 
    \end{tabular}
    } 
    \label{tab:data_source}
\end{table}

\subsection{Additional Qualitative Comparison}
\textbf{Training Datasets:}
As shown in Table~\ref{tab:data_source}, we train both a generative model and a teacher model using seven synthetic datasets to enhance label precision. To effectively mitigate the limitations of synthetic images regarding distribution shift and limited diversity, we generate 20M synthetic data based on the aforementioned datasets, thereby significantly broadening the diversity of the overall samples. Notably, for datasets such as Hypersim, where the original data primarily covers indoor scenes, our generation strategy successfully produces outdoor data largely consistent with depth maps, effectively expanding the scene coverage of the training samples.

\begin{figure}[htbp]
    \centering
\includegraphics[width=\linewidth]{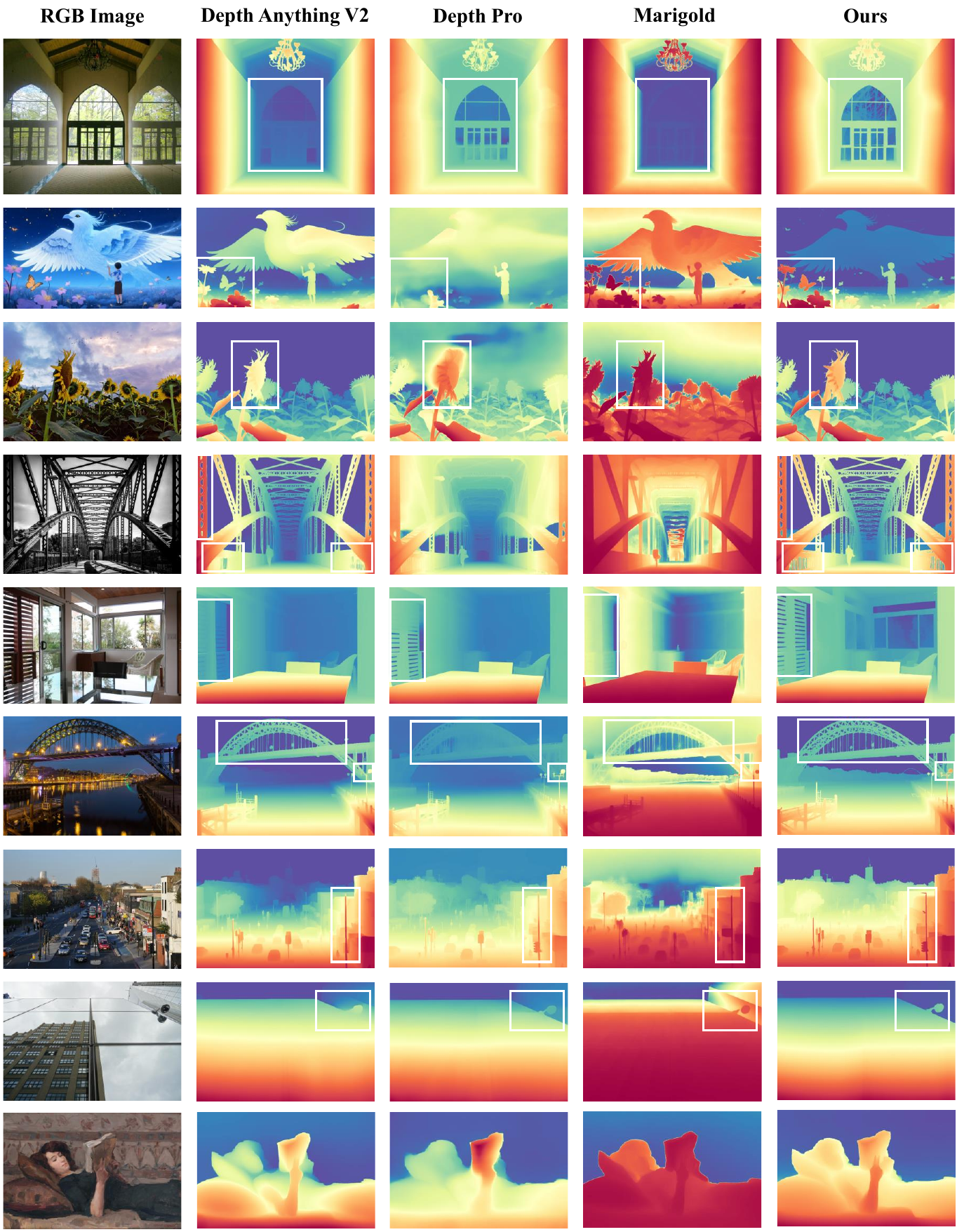}
    \caption{Comparison between our model and leading methods, including Depth Anything V2~\citep{yang2024depth2}, Depth Pro\citep{bochkovskii2024depth} and Marigold\citep{ke2025marigold} on open-world images.}

    \label{fig:case1}
\end{figure}

\begin{figure}[htbp]
    \centering
\includegraphics[width=\linewidth]{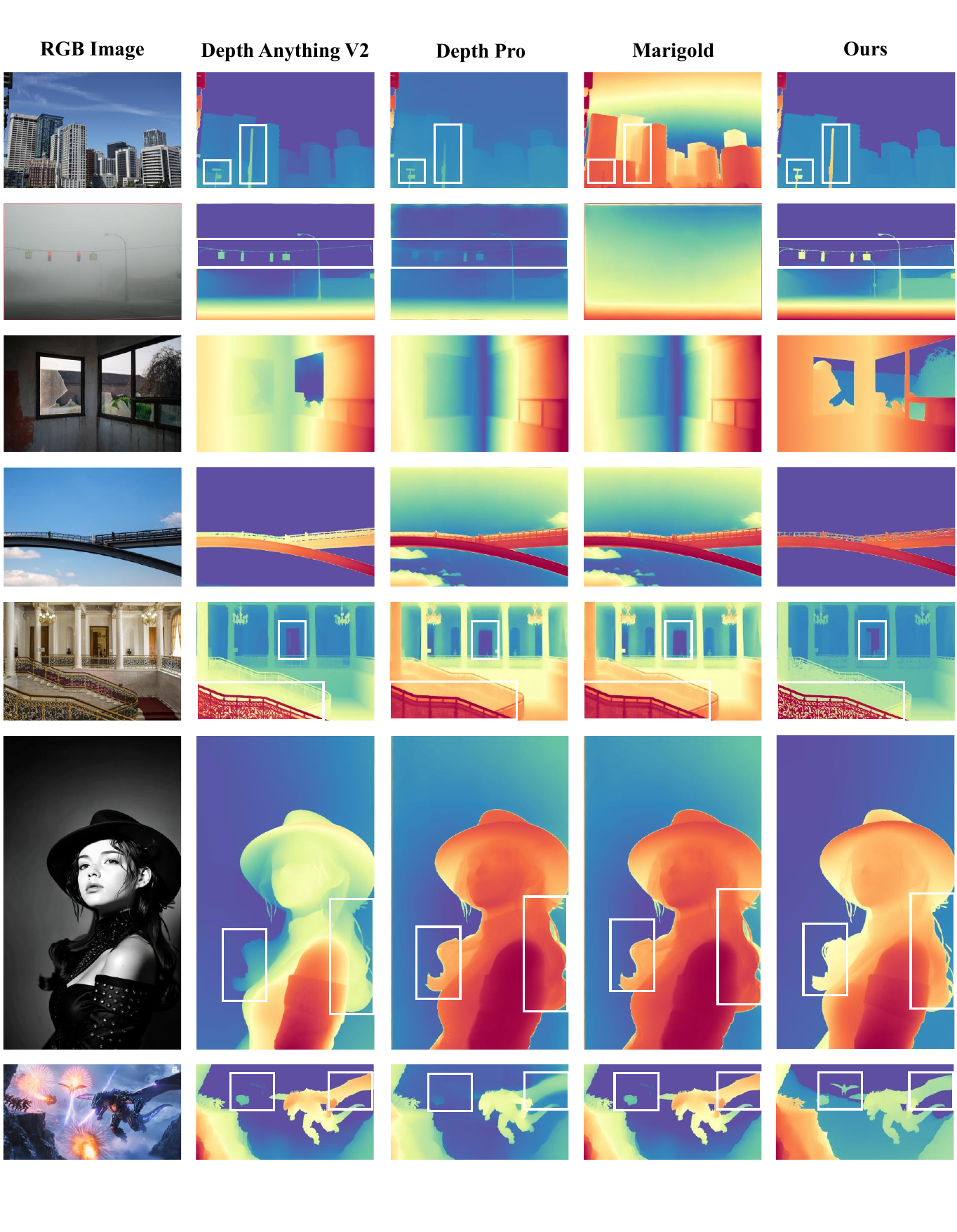}
    \caption{Additional comparison between our model and leading methods, including Depth Anything V2~\citep{yang2024depth2}, Depth Pro\citep{bochkovskii2024depth} and Marigold\citep{ke2025marigold} on ``in-the-wild'' images.
}
    \label{fig:case2}
\end{figure}

\textbf{Qualitative Results:}Figure~\ref{fig:case1} and Figure~\ref{fig:case2} show additional qualitative comparisons with other state-of-the-art monocular depth estimation methods. For indoor, outdoor, and non-real scenes alike, our model consistently produces depth maps with higher fidelity than the other methods. Our method particularly excels in outdoor scenes, showing clear distinctions between distant objects and the sky, while Depth Pro~\citep{bochkovskii2024depth} and Marigold~\citep{ke2025marigold} fail to distinguish objects that are farther away.

\begin{figure}[htbp]
    \centering
\includegraphics[width=\linewidth]{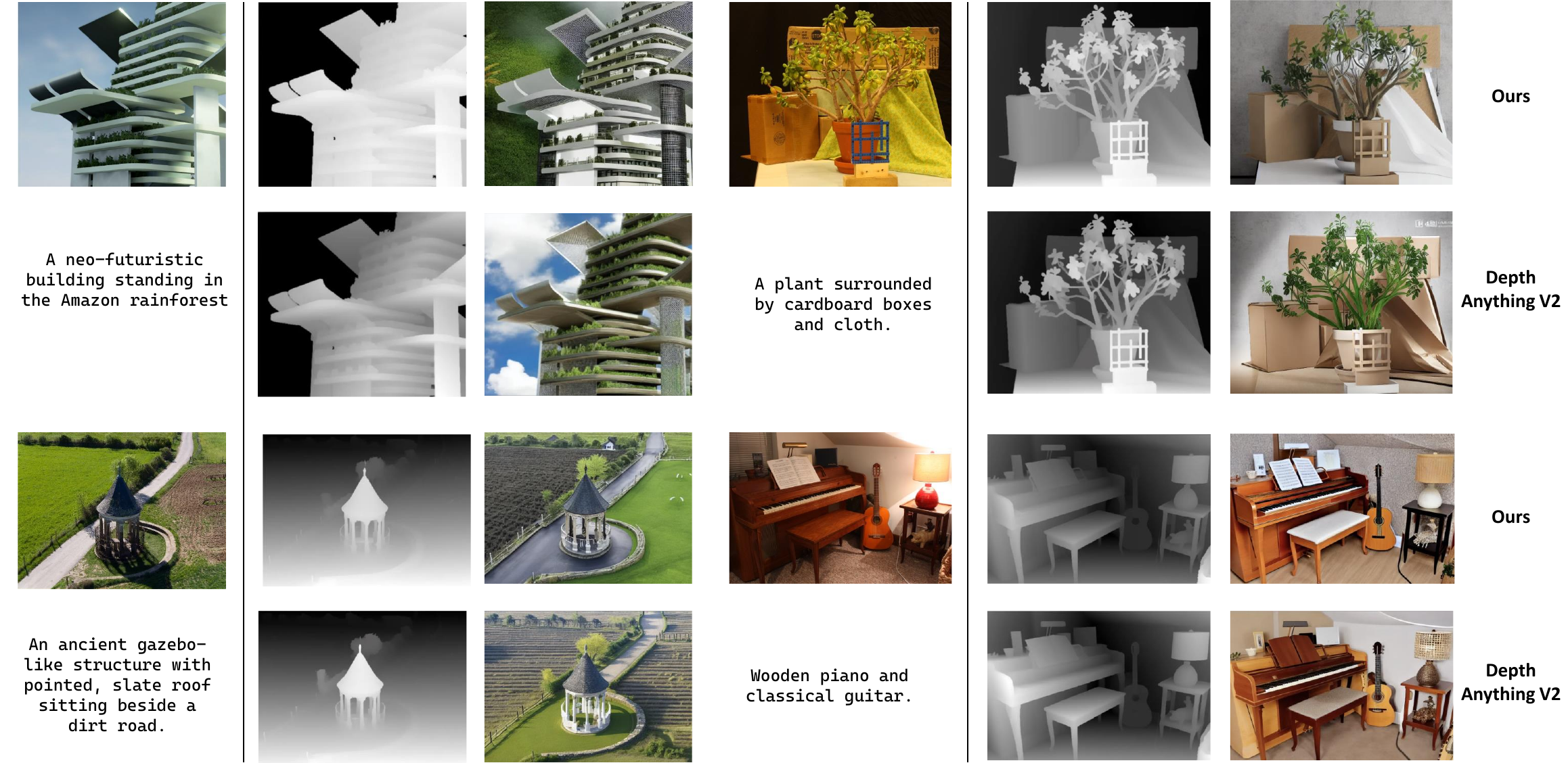}
    \caption{Our model, BRIDGE, generates superior, high-fidelity depth maps that enable ControlNet ~\citep{zhang2023adding} to synthesize new images with a zero-shot capability, precisely replicating the depth field of the source image. In contrast, Depth Anything V2 ~\citep{yang2024depth2} struggles to produce an accurate depth field, as demonstrated by the clear discrepancies between its corresponding ControlNet output and the source images. The prompts used for ControlNet are displayed in the lower left corners, and all images were generated with the same random seed.
}
    \label{fig:control}
\end{figure}

\textbf{Conditional Synthesis:}
Figure~\ref{fig:control} shows a comparison of depth-conditioned synthesis results. We first infer depth based on an image using our \name\ and the Depth Anything V2 model, and then use the most basic pre-trained depth-to-image ControlNet (Zhang, Rao, and Agrawala 2023) with Stable Diffusion 1.5 to synthesize new samples based on the depth maps and a text prompt. We can clearly observe that the depth maps obtained with Depth Anything V2 do not reflect the actual depth well and are additionally inaccurate for some parts of the image. In contrast, our method yields sharp and realistic depth maps. This result is also reflected in the synthesized results, where images created based on our depth map more closely resemble the actual image.

% \subsection{Statement on LLM Usage}
% In the preparation of this manuscript, we utilize Large Language Models (LLMs) only to polish writing.

%% file: main.bbl
\begin{thebibliography}{71}
\providecommand{\natexlab}[1]{#1}
\providecommand{\url}[1]{\texttt{#1}}
\expandafter\ifx\csname urlstyle\endcsname\relax
  \providecommand{\doi}[1]{doi: #1}\else
  \providecommand{\doi}{doi: \begingroup \urlstyle{rm}\Url}\fi

\bibitem[Atapour-Abarghouei \& Breckon(2018)Atapour-Abarghouei and Breckon]{atapour2018real}
Amir Atapour-Abarghouei and Toby~P Breckon.
\newblock Real-time monocular depth estimation using synthetic data with domain adaptation via image style transfer.
\newblock In \emph{Proceedings of the IEEE conference on computer vision and pattern recognition}, pp.\  2800--2810, 2018.

\bibitem[Bhat et~al.(2021)Bhat, Alhashim, and Wonka]{bhat2021adabins}
Shariq~Farooq Bhat, Ibraheem Alhashim, and Peter Wonka.
\newblock Adabins: Depth estimation using adaptive bins.
\newblock In \emph{Proceedings of the IEEE/CVF conference on computer vision and pattern recognition}, pp.\  4009--4018, 2021.

\bibitem[Bhat et~al.(2023)Bhat, Birkl, Wofk, Wonka, and Müller]{zoedepth}
Shariq~Farooq Bhat, Reiner Birkl, Diana Wofk, Peter Wonka, and Matthias Müller.
\newblock Zoedepth: Zero-shot transfer by combining relative and metric depth, 2023.
\newblock URL \url{https://arxiv.org/abs/2302.12288}.

\bibitem[Birkl et~al.(2023)Birkl, Wofk, and M{\"u}ller]{birkl2023midas}
Reiner Birkl, Diana Wofk, and Matthias M{\"u}ller.
\newblock Midas v3. 1--a model zoo for robust monocular relative depth estimation.
\newblock \emph{arXiv preprint arXiv:2307.14460}, 2023.

\bibitem[Bochkovskii et~al.(2024)Bochkovskii, Delaunoy, Germain, Santos, Zhou, Richter, and Koltun]{bochkovskii2024depth}
Aleksei Bochkovskii, Ama{\~A}{\c{G}}l Delaunoy, Hugo Germain, Marcel Santos, Yichao Zhou, Stephan~R Richter, and Vladlen Koltun.
\newblock Depth pro: Sharp monocular metric depth in less than a second.
\newblock \emph{arXiv preprint arXiv:2410.02073}, 2024.

\bibitem[Butler et~al.(2012)Butler, Wulff, Stanley, and Black]{butler2012naturalistic}
Daniel~J Butler, Jonas Wulff, Garrett~B Stanley, and Michael~J Black.
\newblock A naturalistic open source movie for optical flow evaluation.
\newblock In \emph{European conference on computer vision}, pp.\  611--625. Springer, 2012.

\bibitem[Cabon et~al.(2020)Cabon, Murray, and Humenberger]{cabon2020virtual}
Yohann Cabon, Naila Murray, and Martin Humenberger.
\newblock Virtual kitti 2.
\newblock \emph{arXiv preprint arXiv:2001.10773}, 2020.

\bibitem[Charatan et~al.(2024)Charatan, Li, Tagliasacchi, and Sitzmann]{charatan2024pixelsplat}
David Charatan, Sizhe~Lester Li, Andrea Tagliasacchi, and Vincent Sitzmann.
\newblock pixelsplat: 3d gaussian splats from image pairs for scalable generalizable 3d reconstruction.
\newblock In \emph{Proceedings of the IEEE/CVF conference on computer vision and pattern recognition}, pp.\  19457--19467, 2024.

\bibitem[Chen et~al.(2016)Chen, Fu, Yang, and Deng]{chen2016single}
Weifeng Chen, Zhao Fu, Dawei Yang, and Jia Deng.
\newblock Single-image depth perception in the wild.
\newblock \emph{Advances in neural information processing systems}, 29, 2016.

\bibitem[Costanzino et~al.(2023)Costanzino, Ramirez, Poggi, Tosi, Mattoccia, and Di~Stefano]{costanzino2023learning}
Alex Costanzino, Pierluigi~Zama Ramirez, Matteo Poggi, Fabio Tosi, Stefano Mattoccia, and Luigi Di~Stefano.
\newblock Learning depth estimation for transparent and mirror surfaces.
\newblock In \emph{Proceedings of the IEEE/CVF International Conference on Computer Vision}, pp.\  9244--9255, 2023.

\bibitem[Dai et~al.(2017)Dai, Chang, Savva, Halber, Funkhouser, and Nie{\ss}ner]{dai2017scannet}
Angela Dai, Angel~X. Chang, Manolis Savva, Maciej Halber, Thomas Funkhouser, and Matthias Nie{\ss}ner.
\newblock Scannet: Richly-annotated 3d reconstructions of indoor scenes.
\newblock In \emph{Proc. Computer Vision and Pattern Recognition (CVPR), IEEE}, 2017.

\bibitem[Dosovitskiy et~al.(2020)Dosovitskiy, Beyer, Kolesnikov, Weissenborn, Zhai, Unterthiner, Dehghani, Minderer, Heigold, Gelly, et~al.]{dosovitskiy2020image}
Alexey Dosovitskiy, Lucas Beyer, Alexander Kolesnikov, Dirk Weissenborn, Xiaohua Zhai, Thomas Unterthiner, Mostafa Dehghani, Matthias Minderer, Georg Heigold, Sylvain Gelly, et~al.
\newblock An image is worth 16x16 words: Transformers for image recognition at scale.
\newblock \emph{arXiv preprint arXiv:2010.11929}, 2020.

\bibitem[Eigen \& Fergus(2015)Eigen and Fergus]{eigen2015predicting}
David Eigen and Rob Fergus.
\newblock Predicting depth, surface normals and semantic labels with a common multi-scale convolutional architecture.
\newblock In \emph{Proceedings of the IEEE international conference on computer vision}, pp.\  2650--2658, 2015.

\bibitem[Eigen et~al.(2014)Eigen, Puhrsch, and Fergus]{eigen2014depth}
David Eigen, Christian Puhrsch, and Rob Fergus.
\newblock Depth map prediction from a single image using a multi-scale deep network.
\newblock \emph{Advances in neural information processing systems}, 27, 2014.

\bibitem[Fu et~al.(2018)Fu, Gong, Wang, Batmanghelich, and Tao]{fu2018deep}
Huan Fu, Mingming Gong, Chaohui Wang, Kayhan Batmanghelich, and Dacheng Tao.
\newblock Deep ordinal regression network for monocular depth estimation.
\newblock In \emph{Proceedings of the IEEE conference on computer vision and pattern recognition}, pp.\  2002--2011, 2018.

\bibitem[Fu et~al.(2024)Fu, Yin, Hu, Wang, Ma, Tan, Shen, Lin, and Long]{fu2024geowizard}
Xiao Fu, Wei Yin, Mu~Hu, Kaixuan Wang, Yuexin Ma, Ping Tan, Shaojie Shen, Dahua Lin, and Xiaoxiao Long.
\newblock Geowizard: Unleashing the diffusion priors for 3d geometry estimation from a single image.
\newblock In \emph{European Conference on Computer Vision}, pp.\  241--258. Springer, 2024.

\bibitem[Ganin \& Lempitsky(2015)Ganin and Lempitsky]{ganin2015unsupervised}
Yaroslav Ganin and Victor Lempitsky.
\newblock Unsupervised domain adaptation by backpropagation.
\newblock In \emph{International conference on machine learning}, pp.\  1180--1189. PMLR, 2015.

\bibitem[Geiger et~al.(2012)Geiger, Lenz, and Urtasun]{geiger2012we}
Andreas Geiger, Philip Lenz, and Raquel Urtasun.
\newblock Are we ready for autonomous driving? the kitti vision benchmark suite.
\newblock In \emph{2012 IEEE conference on computer vision and pattern recognition}, pp.\  3354--3361. IEEE, 2012.

\bibitem[Godard et~al.(2017)Godard, Mac~Aodha, and Brostow]{godard2017unsupervised}
Cl{\'e}ment Godard, Oisin Mac~Aodha, and Gabriel~J Brostow.
\newblock Unsupervised monocular depth estimation with left-right consistency.
\newblock In \emph{Proceedings of the IEEE conference on computer vision and pattern recognition}, pp.\  270--279, 2017.

\bibitem[Gui et~al.(2025)Gui, Schusterbauer, Prestel, Ma, Kotovenko, Grebenkova, Baumann, Hu, and Ommer]{gui2025depthfm}
Ming Gui, Johannes Schusterbauer, Ulrich Prestel, Pingchuan Ma, Dmytro Kotovenko, Olga Grebenkova, Stefan~Andreas Baumann, Vincent~Tao Hu, and Bj{\"o}rn Ommer.
\newblock Depthfm: Fast generative monocular depth estimation with flow matching.
\newblock In \emph{Proceedings of the AAAI Conference on Artificial Intelligence}, volume~39, pp.\  3203--3211, 2025.

\bibitem[He et~al.(2025)He, Guo, Li, Li, Cui, and Zhang]{he2025distill}
Xiankang He, Dongyan Guo, Hongji Li, Ruibo Li, Ying Cui, and Chi Zhang.
\newblock Distill any depth: Distillation creates a stronger monocular depth estimator.
\newblock \emph{arXiv preprint arXiv:2502.19204}, 2025.

\bibitem[Ho et~al.(2020)Ho, Jain, and Abbeel]{ho2020denoising}
Jonathan Ho, Ajay Jain, and Pieter Abbeel.
\newblock Denoising diffusion probabilistic models.
\newblock \emph{Advances in neural information processing systems}, 33:\penalty0 6840--6851, 2020.

\bibitem[Hu et~al.(2024)Hu, Yin, Zhang, Cai, Long, Chen, Wang, Yu, Shen, and Shen]{hu2024metric3d}
Mu~Hu, Wei Yin, Chi Zhang, Zhipeng Cai, Xiaoxiao Long, Hao Chen, Kaixuan Wang, Gang Yu, Chunhua Shen, and Shaojie Shen.
\newblock Metric3d v2: A versatile monocular geometric foundation model for zero-shot metric depth and surface normal estimation.
\newblock \emph{IEEE Transactions on Pattern Analysis and Machine Intelligence}, 2024.

\bibitem[Ji et~al.(2023)Ji, Chen, Xie, Hong, Liu, Liu, Lu, Li, and Luo]{ji2023ddp}
Yuanfeng Ji, Zhe Chen, Enze Xie, Lanqing Hong, Xihui Liu, Zhaoqiang Liu, Tong Lu, Zhenguo Li, and Ping Luo.
\newblock Ddp: Diffusion model for dense visual prediction.
\newblock In \emph{Proceedings of the IEEE/CVF International Conference on Computer Vision}, pp.\  21741--21752, 2023.

\bibitem[Karami et~al.(2017)Karami, Prasad, and Shehata]{karami2017image}
Ebrahim Karami, Siva Prasad, and Mohamed Shehata.
\newblock Image matching using sift, surf, brief and orb: performance comparison for distorted images.
\newblock \emph{arXiv preprint arXiv:1710.02726}, 2017.

\bibitem[Ke et~al.(2024)Ke, Obukhov, Huang, Metzger, Daudt, and Schindler]{ke2024repurposing}
Bingxin Ke, Anton Obukhov, Shengyu Huang, Nando Metzger, Rodrigo~Caye Daudt, and Konrad Schindler.
\newblock Repurposing diffusion-based image generators for monocular depth estimation.
\newblock In \emph{Proceedings of the IEEE/CVF conference on computer vision and pattern recognition}, pp.\  9492--9502, 2024.

\bibitem[Ke et~al.(2025)Ke, Qu, Wang, Metzger, Huang, Li, Obukhov, and Schindler]{ke2025marigold}
Bingxin Ke, Kevin Qu, Tianfu Wang, Nando Metzger, Shengyu Huang, Bo~Li, Anton Obukhov, and Konrad Schindler.
\newblock Marigold: Affordable adaptation of diffusion-based image generators for image analysis, 2025.

\bibitem[Kerbl et~al.(2023)Kerbl, Kopanas, Leimk{\"u}hler, and Drettakis]{kerbl20233d}
Bernhard Kerbl, Georgios Kopanas, Thomas Leimk{\"u}hler, and George Drettakis.
\newblock 3d gaussian splatting for real-time radiance field rendering.
\newblock \emph{ACM Trans. Graph.}, 42\penalty0 (4):\penalty0 139--1, 2023.

\bibitem[Krizhevsky et~al.(2009)Krizhevsky, Hinton, et~al.]{krizhevsky2009learning}
Alex Krizhevsky, Geoffrey Hinton, et~al.
\newblock Learning multiple layers of features from tiny images.(2009), 2009.

\bibitem[Kuznietsov et~al.(2017)Kuznietsov, Stuckler, and Leibe]{kuznietsov2017semi}
Yevhen Kuznietsov, Jorg Stuckler, and Bastian Leibe.
\newblock Semi-supervised deep learning for monocular depth map prediction.
\newblock In \emph{Proceedings of the IEEE conference on computer vision and pattern recognition}, pp.\  6647--6655, 2017.

\bibitem[Labs(2024)]{flux2024}
Black~Forest Labs.
\newblock Flux.
\newblock \url{https://github.com/black-forest-labs/flux}, 2024.

\bibitem[Lee et~al.(2013)]{lee2013pseudo}
Dong-Hyun Lee et~al.
\newblock Pseudo-label: The simple and efficient semi-supervised learning method for deep neural networks.
\newblock In \emph{Workshop on challenges in representation learning, ICML}, volume~3, pp.\  896. Atlanta, 2013.

\bibitem[Liu et~al.(2021)Liu, Lin, Cao, Hu, Wei, Zhang, Lin, and Guo]{liu2021swin}
Ze~Liu, Yutong Lin, Yue Cao, Han Hu, Yixuan Wei, Zheng Zhang, Stephen Lin, and Baining Guo.
\newblock Swin transformer: Hierarchical vision transformer using shifted windows.
\newblock In \emph{Proceedings of the IEEE/CVF international conference on computer vision}, pp.\  10012--10022, 2021.

\bibitem[Liu et~al.(2022)Liu, Hu, Lin, Yao, Xie, Wei, Ning, Cao, Zhang, Dong, et~al.]{liu2022swin}
Ze~Liu, Han Hu, Yutong Lin, Zhuliang Yao, Zhenda Xie, Yixuan Wei, Jia Ning, Yue Cao, Zheng Zhang, Li~Dong, et~al.
\newblock Swin transformer v2: Scaling up capacity and resolution.
\newblock In \emph{Proceedings of the IEEE/CVF conference on computer vision and pattern recognition}, pp.\  12009--12019, 2022.

\bibitem[Mehl et~al.(2023)Mehl, Schmalfuss, Jahedi, Nalivayko, and Bruhn]{mehl2023spring}
Lukas Mehl, Jenny Schmalfuss, Azin Jahedi, Yaroslava Nalivayko, and Andr{\'e}s Bruhn.
\newblock Spring: A high-resolution high-detail dataset and benchmark for scene flow, optical flow and stereo.
\newblock In \emph{Proceedings of the IEEE/CVF Conference on Computer Vision and Pattern Recognition}, pp.\  4981--4991, 2023.

\bibitem[Mildenhall et~al.(2021)Mildenhall, Srinivasan, Tancik, Barron, Ramamoorthi, and Ng]{mildenhall2021nerf}
Ben Mildenhall, Pratul~P Srinivasan, Matthew Tancik, Jonathan~T Barron, Ravi Ramamoorthi, and Ren Ng.
\newblock Nerf: Representing scenes as neural radiance fields for view synthesis.
\newblock \emph{Communications of the ACM}, 65\penalty0 (1):\penalty0 99--106, 2021.

\bibitem[Nichol \& Dhariwal(2021)Nichol and Dhariwal]{nichol2021improved}
Alexander~Quinn Nichol and Prafulla Dhariwal.
\newblock Improved denoising diffusion probabilistic models.
\newblock In \emph{International conference on machine learning}, pp.\  8162--8171. PMLR, 2021.

\bibitem[Nilsson \& Akenine-M{\"o}ller(2020)Nilsson and Akenine-M{\"o}ller]{nilsson2020understanding}
Jim Nilsson and Tomas Akenine-M{\"o}ller.
\newblock Understanding ssim.
\newblock \emph{arXiv preprint arXiv:2006.13846}, 2020.

\bibitem[Oquab et~al.(2023)Oquab, Darcet, Moutakanni, Vo, Szafraniec, Khalidov, Fernandez, Haziza, Massa, El-Nouby, et~al.]{oquab2023dinov2}
Maxime Oquab, Timoth{\'e}e Darcet, Th{\'e}o Moutakanni, Huy Vo, Marc Szafraniec, Vasil Khalidov, Pierre Fernandez, Daniel Haziza, Francisco Massa, Alaaeldin El-Nouby, et~al.
\newblock Dinov2: Learning robust visual features without supervision.
\newblock \emph{arXiv preprint arXiv:2304.07193}, 2023.

\bibitem[Patil et~al.(2022)Patil, Sakaridis, Liniger, and Van~Gool]{patil2022p3depth}
Vaishakh Patil, Christos Sakaridis, Alexander Liniger, and Luc Van~Gool.
\newblock P3depth: Monocular depth estimation with a piecewise planarity prior.
\newblock In \emph{Proceedings of the IEEE/CVF Conference on Computer Vision and Pattern Recognition}, pp.\  1610--1621, 2022.

\bibitem[Piccinelli et~al.(2024)Piccinelli, Yang, Sakaridis, Segu, Li, Van~Gool, and Yu]{piccinelli2024unidepth}
Luigi Piccinelli, Yung-Hsu Yang, Christos Sakaridis, Mattia Segu, Siyuan Li, Luc Van~Gool, and Fisher Yu.
\newblock Unidepth: Universal monocular metric depth estimation.
\newblock In \emph{Proceedings of the IEEE/CVF Conference on Computer Vision and Pattern Recognition}, pp.\  10106--10116, 2024.

\bibitem[Prabhudesai et~al.(2024)Prabhudesai, Mendonca, Qin, Fragkiadaki, and Pathak]{prabhudesai2024video}
Mihir Prabhudesai, Russell Mendonca, Zheyang Qin, Katerina Fragkiadaki, and Deepak Pathak.
\newblock Video diffusion alignment via reward gradients.
\newblock \emph{arXiv preprint arXiv:2407.08737}, 2024.

\bibitem[Rajpal et~al.(2023)Rajpal, Cheema, Illgner-Fehns, Slusallek, and Jaiswal]{rajpal2023high}
Aakash Rajpal, Noshaba Cheema, Klaus Illgner-Fehns, Philipp Slusallek, and Sunil Jaiswal.
\newblock High-resolution synthetic rgb-d datasets for monocular depth estimation.
\newblock In \emph{Proceedings of the IEEE/CVF conference on computer vision and pattern recognition}, pp.\  1188--1198, 2023.

\bibitem[Ranftl et~al.(2020)Ranftl, Lasinger, Hafner, Schindler, and Koltun]{ranftl2020towards}
Ren{\'e} Ranftl, Katrin Lasinger, David Hafner, Konrad Schindler, and Vladlen Koltun.
\newblock Towards robust monocular depth estimation: Mixing datasets for zero-shot cross-dataset transfer.
\newblock \emph{IEEE transactions on pattern analysis and machine intelligence}, 44\penalty0 (3):\penalty0 1623--1637, 2020.

\bibitem[Ranftl et~al.(2021)Ranftl, Bochkovskiy, and Koltun]{ranftl2021vision}
Ren{\'e} Ranftl, Alexey Bochkovskiy, and Vladlen Koltun.
\newblock Vision transformers for dense prediction.
\newblock In \emph{Proceedings of the IEEE/CVF international conference on computer vision}, pp.\  12179--12188, 2021.

\bibitem[Rasla \& Beyeler(2022)Rasla and Beyeler]{rasla2022relative}
Alex Rasla and Michael Beyeler.
\newblock The relative importance of depth cues and semantic edges for indoor mobility using simulated prosthetic vision in immersive virtual reality.
\newblock In \emph{Proceedings of the 28th ACM symposium on virtual reality software and technology}, pp.\  1--11, 2022.

\bibitem[Ren et~al.(2020)Ren, Raj, El-Khamy, and Lee]{ren2020suw}
Haoyu Ren, Aman Raj, Mostafa El-Khamy, and Jungwon Lee.
\newblock Suw-learn: Joint supervised, unsupervised, weakly supervised deep learning for monocular depth estimation.
\newblock In \emph{Proceedings of the IEEE/CVF Conference on Computer Vision and Pattern Recognition Workshops}, pp.\  750--751, 2020.

\bibitem[Roberts et~al.(2021)Roberts, Ramapuram, Ranjan, Kumar, Bautista, Paczan, Webb, and Susskind]{roberts2021hypersim}
Mike Roberts, Jason Ramapuram, Anurag Ranjan, Atulit Kumar, Miguel~Angel Bautista, Nathan Paczan, Russ Webb, and Joshua~M Susskind.
\newblock Hypersim: A photorealistic synthetic dataset for holistic indoor scene understanding.
\newblock In \emph{Proceedings of the IEEE/CVF international conference on computer vision}, pp.\  10912--10922, 2021.

\bibitem[Saxena et~al.(2008)Saxena, Sun, and Ng]{saxena2008make3d}
Ashutosh Saxena, Min Sun, and Andrew~Y Ng.
\newblock Make3d: Learning 3d scene structure from a single still image.
\newblock \emph{IEEE transactions on pattern analysis and machine intelligence}, 31\penalty0 (5):\penalty0 824--840, 2008.

\bibitem[Schops et~al.(2017)Schops, Schonberger, Galliani, Sattler, Schindler, Pollefeys, and Geiger]{schops2017multi}
Thomas Schops, Johannes~L Schonberger, Silvano Galliani, Torsten Sattler, Konrad Schindler, Marc Pollefeys, and Andreas Geiger.
\newblock A multi-view stereo benchmark with high-resolution images and multi-camera videos.
\newblock In \emph{Proceedings of the IEEE conference on computer vision and pattern recognition}, pp.\  3260--3269, 2017.

\bibitem[Shao et~al.(2023{\natexlab{a}})Shao, Pei, Chen, Wu, and Li]{shao2023nddepth}
Shuwei Shao, Zhongcai Pei, Weihai Chen, Xingming Wu, and Zhengguo Li.
\newblock Nddepth: Normal-distance assisted monocular depth estimation.
\newblock In \emph{Proceedings of the IEEE/CVF International Conference on Computer Vision}, pp.\  7931--7940, 2023{\natexlab{a}}.

\bibitem[Shao et~al.(2023{\natexlab{b}})Shao, Pei, Wu, Liu, Chen, and Li]{shao2023iebins}
Shuwei Shao, Zhongcai Pei, Xingming Wu, Zhong Liu, Weihai Chen, and Zhengguo Li.
\newblock Iebins: Iterative elastic bins for monocular depth estimation.
\newblock \emph{Advances in Neural Information Processing Systems}, 36:\penalty0 53025--53037, 2023{\natexlab{b}}.

\bibitem[Silberman et~al.(2012)Silberman, Hoiem, Kohli, and Fergus]{silberman2012indoor}
Nathan Silberman, Derek Hoiem, Pushmeet Kohli, and Rob Fergus.
\newblock Indoor segmentation and support inference from rgbd images.
\newblock In \emph{European conference on computer vision}, pp.\  746--760. Springer, 2012.

\bibitem[Sohn et~al.(2020)Sohn, Berthelot, Carlini, Zhang, Zhang, Raffel, Cubuk, Kurakin, and Li]{sohn2020fixmatch}
Kihyuk Sohn, David Berthelot, Nicholas Carlini, Zizhao Zhang, Han Zhang, Colin~A Raffel, Ekin~Dogus Cubuk, Alexey Kurakin, and Chun-Liang Li.
\newblock Fixmatch: Simplifying semi-supervised learning with consistency and confidence.
\newblock \emph{Advances in neural information processing systems}, 33:\penalty0 596--608, 2020.

\bibitem[Song et~al.(2020)Song, Meng, and Ermon]{song2020denoising}
Jiaming Song, Chenlin Meng, and Stefano Ermon.
\newblock Denoising diffusion implicit models.
\newblock \emph{arXiv preprint arXiv:2010.02502}, 2020.

\bibitem[Wang et~al.(2019{\natexlab{a}})Wang, Zheng, Yan, Deng, Zhao, and Chu]{wang2019irs}
Qiang Wang, Shizhen Zheng, Qingsong Yan, Fei Deng, Kaiyong Zhao, and Xiaowen Chu.
\newblock Irs: A large naturalistic indoor robotics stereo dataset to train deep models for disparity and surface normal estimation.
\newblock \emph{arXiv preprint arXiv:1912.09678}, 2019{\natexlab{a}}.

\bibitem[Wang et~al.(2025)Wang, Xu, Dai, Xiang, Deng, Tong, and Yang]{wang2025moge}
Ruicheng Wang, Sicheng Xu, Cassie Dai, Jianfeng Xiang, Yu~Deng, Xin Tong, and Jiaolong Yang.
\newblock Moge: Unlocking accurate monocular geometry estimation for open-domain images with optimal training supervision.
\newblock In \emph{Proceedings of the Computer Vision and Pattern Recognition Conference}, pp.\  5261--5271, 2025.

\bibitem[Wang et~al.(2020)Wang, Zhu, Wang, Hu, Qiu, Wang, Hu, Kapoor, and Scherer]{wang2020tartanair}
Wenshan Wang, Delong Zhu, Xiangwei Wang, Yaoyu Hu, Yuheng Qiu, Chen Wang, Yafei Hu, Ashish Kapoor, and Sebastian Scherer.
\newblock Tartanair: A dataset to push the limits of visual slam.
\newblock In \emph{2020 IEEE/RSJ International Conference on Intelligent Robots and Systems (IROS)}, pp.\  4909--4916. IEEE, 2020.

\bibitem[Wang et~al.(2019{\natexlab{b}})Wang, Chao, Garg, Hariharan, Campbell, and Weinberger]{wang2019pseudo}
Yan Wang, Wei-Lun Chao, Divyansh Garg, Bharath Hariharan, Mark Campbell, and Kilian~Q Weinberger.
\newblock Pseudo-lidar from visual depth estimation: Bridging the gap in 3d object detection for autonomous driving.
\newblock In \emph{Proceedings of the IEEE/CVF conference on computer vision and pattern recognition}, pp.\  8445--8453, 2019{\natexlab{b}}.

\bibitem[Wofk et~al.(2019)Wofk, Ma, Yang, Karaman, and Sze]{wofk2019fastdepth}
Diana Wofk, Fangchang Ma, Tien-Ju Yang, Sertac Karaman, and Vivienne Sze.
\newblock Fastdepth: Fast monocular depth estimation on embedded systems.
\newblock In \emph{2019 International Conference on Robotics and Automation (ICRA)}, pp.\  6101--6108. IEEE, 2019.

\bibitem[Xie et~al.(2020)Xie, Dai, Hovy, Luong, and Le]{xie2020unsupervised}
Qizhe Xie, Zihang Dai, Eduard Hovy, Thang Luong, and Quoc Le.
\newblock Unsupervised data augmentation for consistency training.
\newblock \emph{Advances in neural information processing systems}, 33:\penalty0 6256--6268, 2020.

\bibitem[Yang et~al.(2024{\natexlab{a}})Yang, Huang, Yin, Shen, Liu, He, Lin, Ouyang, and He]{yang2024depth}
Honghui Yang, Di~Huang, Wei Yin, Chunhua Shen, Haifeng Liu, Xiaofei He, Binbin Lin, Wanli Ouyang, and Tong He.
\newblock Depth any video with scalable synthetic data.
\newblock \emph{arXiv preprint arXiv:2410.10815}, 2024{\natexlab{a}}.

\bibitem[Yang et~al.(2024{\natexlab{b}})Yang, Huang, Yin, Shen, Liu, He, Lin, Ouyang, and He]{yang2024depthvideo}
Honghui Yang, Di~Huang, Wei Yin, Chunhua Shen, Haifeng Liu, Xiaofei He, Binbin Lin, Wanli Ouyang, and Tong He.
\newblock Depth any video with scalable synthetic data.
\newblock \emph{arXiv preprint arXiv:2410.10815}, 2024{\natexlab{b}}.

\bibitem[Yang et~al.(2023{\natexlab{a}})Yang, Zhao, Qi, Qiao, Shi, and Zhao]{yang2023shrinking}
Lihe Yang, Zhen Zhao, Lei Qi, Yu~Qiao, Yinghuan Shi, and Hengshuang Zhao.
\newblock Shrinking class space for enhanced certainty in semi-supervised learning.
\newblock In \emph{Proceedings of the IEEE/CVF international conference on computer vision}, pp.\  16187--16196, 2023{\natexlab{a}}.

\bibitem[Yang et~al.(2024{\natexlab{c}})Yang, Kang, Huang, Xu, Feng, and Zhao]{yang2024depth1}
Lihe Yang, Bingyi Kang, Zilong Huang, Xiaogang Xu, Jiashi Feng, and Hengshuang Zhao.
\newblock Depth anything: Unleashing the power of large-scale unlabeled data.
\newblock In \emph{Proceedings of the IEEE/CVF conference on computer vision and pattern recognition}, pp.\  10371--10381, 2024{\natexlab{c}}.

\bibitem[Yang et~al.(2024{\natexlab{d}})Yang, Kang, Huang, Zhao, Xu, Feng, and Zhao]{yang2024depth2}
Lihe Yang, Bingyi Kang, Zilong Huang, Zhen Zhao, Xiaogang Xu, Jiashi Feng, and Hengshuang Zhao.
\newblock Depth anything v2.
\newblock \emph{Advances in Neural Information Processing Systems}, 37:\penalty0 21875--21911, 2024{\natexlab{d}}.

\bibitem[Yang et~al.(2023{\natexlab{b}})Yang, Ma, Ji, and Ren]{yang2023gedepth}
Xiaodong Yang, Zhuang Ma, Zhiyu Ji, and Zhe Ren.
\newblock Gedepth: Ground embedding for monocular depth estimation.
\newblock In \emph{Proceedings of the IEEE/CVF international conference on computer vision}, pp.\  12719--12727, 2023{\natexlab{b}}.

\bibitem[Yao et~al.(2020)Yao, Luo, Li, Zhang, Ren, Zhou, Fang, and Quan]{yao2020blendedmvs}
Yao Yao, Zixin Luo, Shiwei Li, Jingyang Zhang, Yufan Ren, Lei Zhou, Tian Fang, and Long Quan.
\newblock Blendedmvs: A large-scale dataset for generalized multi-view stereo networks.
\newblock In \emph{Proceedings of the IEEE/CVF conference on computer vision and pattern recognition}, pp.\  1790--1799, 2020.

\bibitem[Ye et~al.(2024)Ye, Nie, Chang, Chen, Zhi, and Han]{ye2024gaustudio}
Chongjie Ye, Yinyu Nie, Jiahao Chang, Yuantao Chen, Yihao Zhi, and Xiaoguang Han.
\newblock Gaustudio: A modular framework for 3d gaussian splatting and beyond.
\newblock \emph{arXiv preprint arXiv:2403.19632}, 2024.

\bibitem[Yin et~al.(2021)Yin, Zhang, Wang, Niklaus, Mai, Chen, and Shen]{yin2021learning}
Wei Yin, Jianming Zhang, Oliver Wang, Simon Niklaus, Long Mai, Simon Chen, and Chunhua Shen.
\newblock Learning to recover 3d scene shape from a single image.
\newblock In \emph{Proceedings of the IEEE/CVF conference on computer vision and pattern recognition}, pp.\  204--213, 2021.

\bibitem[Zhang et~al.(2023)Zhang, Rao, and Agrawala]{zhang2023adding}
Lvmin Zhang, Anyi Rao, and Maneesh Agrawala.
\newblock Adding conditional control to text-to-image diffusion models.
\newblock In \emph{Proceedings of the IEEE/CVF international conference on computer vision}, pp.\  3836--3847, 2023.

\end{thebibliography}
